\documentclass[letterpaper]{article}

\usepackage[T1]{fontenc}
\usepackage{geometry}
\usepackage{setspace}
\usepackage[style = chem-acs]{biblatex}
\usepackage{graphicx}
\usepackage{float}
\usepackage{booktabs}
\usepackage{multirow}
\usepackage{amsmath}
\usepackage{amssymb}
\usepackage{xcolor}
\usepackage{hyperref}
\usepackage[labelfont=bf]{caption}
\usepackage{subcaption}

\newfloat{scheme}{htbp}{los}
\floatname{scheme}{Scheme}

\newcommand{\mocop}{\textsc{MoCoP}}

\usepackage{authblk}
\author[1]{Jie Li\thanks{Corresponding author: jerry.8.li@gsk.com}}
\author[2]{Kathryn E. Kirchoff}
\author[2]{Dante A. Pertusi}
\author[1]{Zhizhuo Zhang}
\affil[1]{GSK, Artificial Intelligence and Machine Learning}
\affil[2]{GSK, Cheminformatics}

\title{Improving Molecular-Morphology Contrastive Pretraining\\
using Deep-Learning-based Morphology Profiles}

\date{}

\begin{document}

\maketitle

\begin{abstract}
Recent advancements in image-based profiling techniques have enabled the collection of high-volume cell morphology data, allowing new molecular embedding models to learn from the experimental phenotypic perturbations of a molecule in a cell.
Previously, we developed Molecule-Morphology Contrastive Pretraining (\mocop{}), a strategy for aligning small molecule embeddings to morphology fingerprints extracted through CellProfiler.
The resulting molecular representation showed transferable performance for quantitative structure--activity relationship (QSAR) prediction tasks.
Here, we extend the method by using a deep-learning-based cell image encoding pipeline to extract more feature-rich morphology profiles and align them to the molecular embeddings through contrastive learning.
The new embeddings encode more accurate information on how molecules perturb cell morphology and enable improvements for QSAR predictions through either fixed-embedding linear probes or fully flexible fine-tuning.
Morphology retrieval performance scales log-linearly with training data size, suggesting continued improvements as larger datasets become available.
The improved \mocop{} v2 also achieves superior performance on toxicity prediction and competitive results on ADME and activity benchmarks, when compared with existing molecular embedding models that use both cell morphology and transcriptomic data during training.
\end{abstract}

\section{Introduction}

The identification of molecular representations that encode biologically relevant information is a central challenge in computational drug discovery.
Traditional molecular descriptors such as extended-connectivity fingerprints (ECFPs)~\cite{Rogers2010ECFP} capture substructural motifs but are inherently limited to information derivable from chemical graphs alone.
Recent progress in deep learning has produced a rich landscape of learned molecular representations, including graph neural networks (GNNs) pretrained with self-supervised objectives~\cite{Hu2019PretrainingGNN, Rong2020GROVER, Liu2021GraphMVP, Xia2023MoleBERT, Sun2021MoCL, Wang2021MolCLR} and transformer-based models operating on SMILES strings~\cite{Ross2021MoLFormer, Ahmad2022ChemBERTa2, Fabian2020MolBert}.
Three-dimensional molecular structure has also been incorporated through frameworks such as Uni-Mol~\cite{Zhou2023UniMol}, Newtonnet~\cite{newtonnet} and PaiNN~\cite{painn}.
While these methods learn expressive representations from chemical data, they do not directly capture how a molecule interacts with the complex machinery of a living cell.

In parallel, image-based profiling, particularly the Cell Painting assay~\cite{Bray2017CellPainting, Gustafsdottir2013CellPainting}, has emerged as a scalable technique for measuring cellular responses to chemical perturbations.
By staining six cellular compartments and acquiring multiplexed fluorescence images, Cell Painting yields high-dimensional morphological readouts that capture drug-induced phenotypic changes~\cite{Way2020CellHealth, Way2022Complementary}.
Large-scale joint efforts such as the Joint Undertaking for Morphological Profiling (JUMP) Cell Painting Consortium have generated image datasets spanning over 116,000 compounds~\cite{Chandrasekaran2023JUMPCP, Chandrasekaran2022CPJUMP1}, creating an unprecedented resource for studying chemical biology at scale.
Several studies have demonstrated the utility of these morphological profiles for predicting compound bioactivity~\cite{Becker2020PredictingActivity, Hofmarcher2019CNN, Simm2018Repurposing, Haslum2023Bioactivity}, mechanism of action~\cite{Trapotsi2021Comparison, Tian2022CombiningMolCP}, and drug targets~\cite{Iyer2025CellMorphGenes}.

Motivated by the complementary nature of chemical structure and cellular morphology, a growing body of work has explored cross-modal contrastive learning to align molecular and phenotypic representations.
CLOOME~\cite{SanchezFernandez2023CLOOME} adapted the CLIP framework~\cite{Radford2021CLIP} to learn joint embeddings of chemical structures and Cell Painting images.
More recently, InfoAlign~\cite{Liu2024InfoAlign} introduced an information-theoretic approach that integrates molecules with both morphological and transcriptomic data through a context graph, achieving superior performance on multiple downstream benchmarks.
CHMR~\cite{Li2025CHMR} further improved upon InfoAlign by modeling hierarchical dependencies across molecular, cellular, and genomic levels, achieveing state-of-the-art performance on multiple biological assay readout predictions, and absorption, distribution, metabolism, excretion and toxicity (ADMET) prediction.
Other notable approaches include MolPhenix~\cite{Fradkin2024MolPhenix}, which demonstrated improved retrieval through pre-trained phenomics models; PhenoScreen~\cite{Wang2024PhenoScreen}, which employed dual-space contrastive learning; MINER~\cite{Rao2025MINER}, which addressed negative sampling calibration; and CellCLIP~\cite{Lu2025CellCLIP}, which leveraged text-guided contrastive learning.
Cross-modal strategies involving transcriptomics have also been explored~\cite{Ha2025CrossModality, Bendidi2025CrossModal, Finlayson2019CrossModal}.

Despite these advances, a critical bottleneck in molecule--morphology alignment lies in the quality of the morphological representations themselves.
The conventional approach relies on CellProfiler~\cite{McQuin2018CellProfiler}, a rule-based image analysis pipeline that extracts handcrafted features such as intensity statistics, texture, and shape descriptors.
While CellProfiler features are interpretable and widely used, they may miss subtle morphological patterns that are difficult to capture with predefined feature extractors.
Recent work in self-supervised representation learning for microscopy has shown that deep learning models, including DINO~\cite{Doron2023DINO, CrossZamirski2022WSDINO}, masked autoencoders~\cite{Kraus2024MAE, Kraus2023MAE}, and supervised models~\cite{Moshkov2022CellPaintingCNN, Kim2024SSLCellPainting}, can learn cellular representations that outperform CellProfiler on downstream tasks such as mechanism-of-action prediction and batch effect correction~\cite{Arevalo2024BatchCorrection, Lin2022BEN}.
This suggests that replacing CellProfiler features with learned morphological embeddings could substantially improve molecule--morphology contrastive pretraining.

Previously, we developed Molecule-Morphology Contrastive Pretraining (\mocop{})~\cite{Nguyen2023MoCoP}, which aligned molecular GNN embeddings with CellProfiler-derived morphological profiles using data from the JUMP-CP Consortium.
\mocop{} v1 demonstrated that the resulting molecular representations consistently improved GNN performance on quantitative structure--activity relationship (QSAR) prediction tasks across varying dataset sizes.
In this work, we introduce \mocop{} v2, which replaces CellProfiler features with deep-learning (DL) morphological embeddings as the alignment target.
These DL embeddings are extracted from Cell Painting images of the JUMP-CP Consortium using a fine-tuned convolutional neural network, followed by CORAL-based batch correction~\cite{Sun2016CORAL} to mitigate inter-source variability.
We show that this upgrade produces molecular representations that (1) more accurately reflect how molecules perturb cell morphology, (2) improve QSAR prediction on ChEMBL20 benchmarks, (3) exhibit log-linear scaling of retrieval performance with training data size, and (4) achieve competitive or state-of-the-art performance on downstream toxicity and ADME prediction tasks compared to recent methods including InfoAlign and CHMR, despite using simpler architecture, and only using cell morphology without transcriptomic data.

\section{Methods}

\begin{figure}[htbp]
  \centering
  \includegraphics[width=\textwidth]{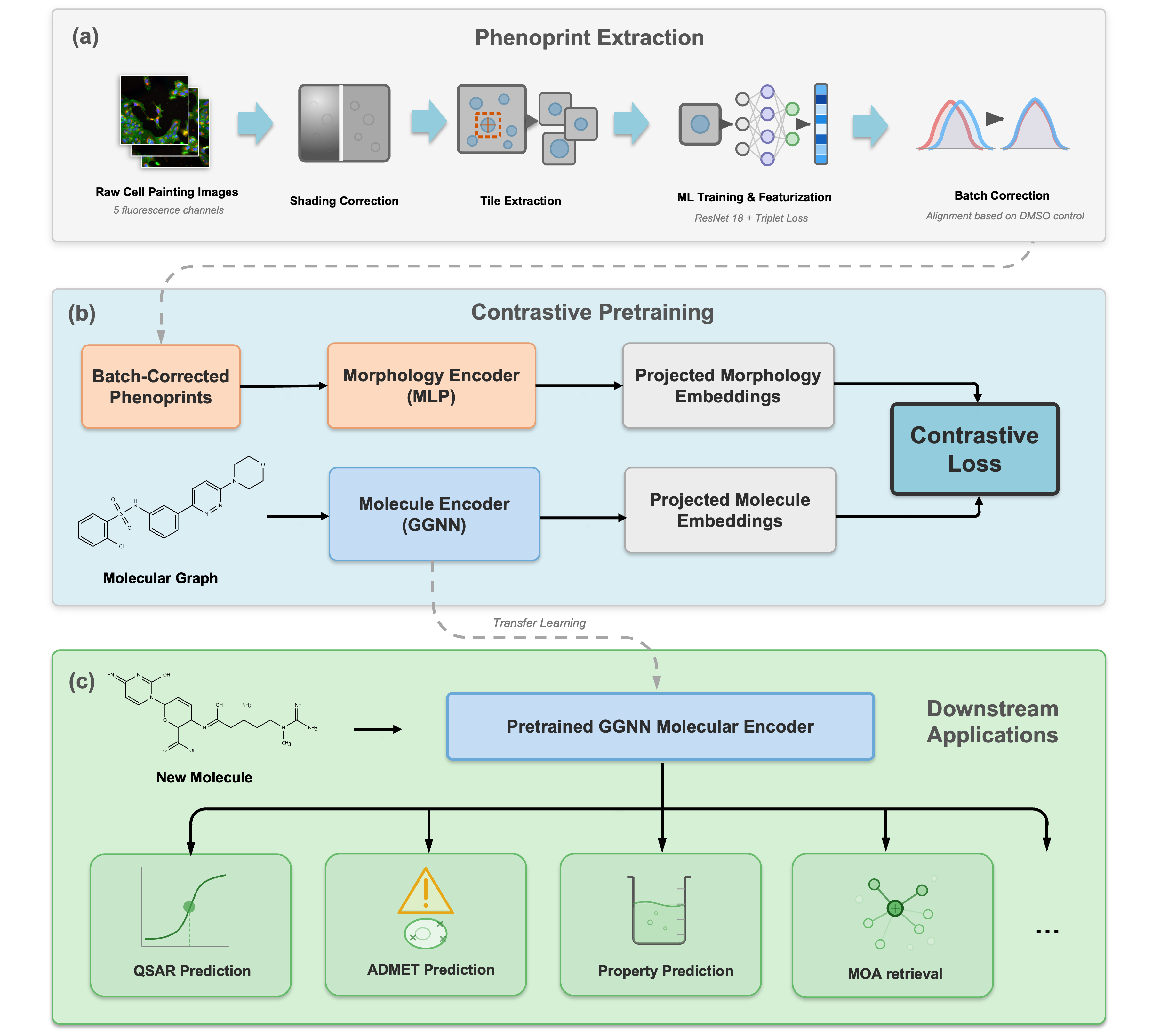}
  \caption{Overview of the \mocop{} v2 pipeline.
  \textbf{(a)}~DL embedding extraction: raw 5-channel Cell Painting images undergo shading correction and tile extraction ($224 \times 224$ px), followed by training and featurization with a ResNet-18 model fine-tuned using triplet loss. The resulting tile-level embeddings are mean-aggregated per well to produce 128-dimensional DL morphological embeddings, which are then batch-corrected using CORAL alignment based on DMSO negative controls.
  \textbf{(b)}~Contrastive pretraining: batch-corrected DL embeddings are encoded by a morphology multi-layer perceptron network (MLP), while molecular graphs are encoded by a gated graph neural network (GGNN). Both branches are projected into a shared 128-dimensional space, and the encoders are jointly trained with a symmetric InfoNCE contrastive loss.
  \textbf{(c)}~Downstream applications: the pretrained GGNN molecular encoder is transferred to new molecules for QSAR prediction, ADMET prediction, molecular property prediction, mechanism-of-action retrieval, and other tasks.}
  \label{fig:pipeline}
\end{figure}

The \mocop{} v2 framework consists of three stages (Figure~\ref{fig:pipeline}): (1)~training a deep-learning image encoder on Cell Painting images to extract DL morphological embeddings, including shading correction, tile extraction, and CORAL batch correction; (2)~molecule--morphology contrastive pretraining, in which a molecular GNN encoder is aligned to the DL embedding space while the image encoder remains frozen; and (3)~transfer of the pretrained molecular encoder to downstream applications.

\subsection{Deep-Learning-based Morphological Embeddings}

The primary methodological advance in \mocop{} v2 is the replacement of CellProfiler-derived morphological features with deep-learning-based (DL) embeddings.
Whereas \mocop{} v1 used CellProfiler to extract handcrafted features (intensity, texture, shape descriptors) from Cell Painting images, DL embeddings are learned representations obtained by fine-tuning a convolutional neural network to distinguish compound treatments from their morphological effects.
The DL embedding extraction pipeline proceeds in three stages: image preprocessing, model training, and inference.

\textbf{Image Preprocessing.}
Raw Cell Painting images are first corrected for uneven illumination using an illumination correction function (ICF) estimated per channel for each plate.
The ICF is computed by smoothing the 10th-percentile intensity image with a Gaussian filter ($\sigma = 50$, kernel size $= 250$ pixels), and each raw image is divided by this correction field.
Next, individual cells are localized using a Laplacian-of-Gaussian (LoG) nuclei detector with sigma parameters scaled by the microscope magnification (at $20\times$: $\sigma_{\min} = 10$, $\sigma_{\max} = 20$ pixels), and a $224 \times 224$ pixel tile is extracted centered on each detected nucleus, with overlapping tiles filtered to avoid redundancy.
All five fluorescence channels (DNA, endoplasmic reticulum, actin/Golgi/plasma membrane, mitochondria, and nucleoli/RNA) are bundled into each tile and stored as HDF5 files for efficient data loading.

\textbf{Model Architecture and Training.}
We employed a ResNet-18~\cite{He2016ResNet} backbone pretrained on ImageNet.
Since the standard model expects three-channel RGB input, the first convolutional layer (\texttt{conv1}) was replaced with a new layer accepting five input channels while preserving the original 64 output feature maps, $7 \times 7$ kernel size, stride of 2, and padding of 3; pretrained weights for all subsequent layers were retained.
The backbone feeds into an MLP projection head with hidden dimensions of 1024 and 128, producing 128-dimensional tile-level embeddings.
During both training and inference, 12 tiles are randomly sampled per well, passed through the network independently, and mean-aggregated to produce a single well-level DL embedding.

The model was fine-tuned using a triplet margin loss with online semihard mining~\cite{Schroff2015FaceNet}.
For each valid triplet $(z_a, z_p, z_n)$, where $z_a$ is an anchor, $z_p$ a positive (same compound), and $z_n$ a negative (different compound), the per-triplet loss is:
\begin{equation}
  \ell(z_a, z_p, z_n) = \max\bigl(0,\; \|z_a - z_p\|_2 - \|z_a - z_n\|_2 + m\bigr)
  \label{eq:triplet}
\end{equation}
where $m = 0.1$ is the margin.
Semihard mining retains only triplets where the negative is farther than the positive but within the margin ($0 < \ell < m$), and the final loss is the mean over all such triplets in the batch; embeddings are not L2-normalized before distance computation.
A balanced sampler ensured each training batch contained 8 distinct compound classes with $2\times$ oversampling to handle class imbalance.
The model was trained on all 217 plates from Source~3 of the JUMP-CP Consortium using AdamW~\cite{Loshchilov2019AdamW} for 100 epochs on 4 NVIDIA A100 80\,GB GPUs, with 1-nearest-neighbor Euclidean accuracy as the primary evaluation metric.
Source~3 was selected for image encoder training because it is the largest single data source in the JUMP-CP dataset, providing 54,655 wells across 217 plates with consistent imaging conditions from a single laboratory.
Training the image encoder on a single source avoids conflating morphological variation with batch effects during the supervised triplet learning stage; inter-source variability is instead addressed downstream through CORAL batch correction.
More details about training can be found in \ref{app:dl_emb_hparams}.

\textbf{Inference.}
After training, the model was applied to extract DL embeddings for every well across all 1,728 plates in the full JUMP-CP dataset, spanning multiple data sources and experimental batches.
Importantly, the image encoder is trained once and then frozen: the resulting DL embeddings are pre-extracted and stored as fixed 128-dimensional vectors.
During the subsequent molecule--morphology contrastive pretraining, the DL embeddings serve as static input to a trainable morphology projection network; the image encoder itself is not revisited.

\subsection{Batch Correction with CORAL}
\label{sec:coral}

Cell Painting experiments conducted across different laboratories, instruments, and time points introduce substantial batch effects that can dominate the embedding space, causing representations to cluster by data source rather than by biological perturbation~\cite{Arevalo2024BatchCorrection, Sypetkowski2023RxRx1}.
To address this, we applied CORrelation ALignment (CORAL)~\cite{Sun2016CORAL} to correct batch effects in the extracted DL embeddings. This batch correction step is essential: without it, the DL embeddings are dominated by data-source identity rather than compound-induced morphological changes (see \ref{app:coral} for visualization).

CORAL aligns the second-order statistics (covariance matrices) of embeddings from different batches.
Specifically, let $C_s$ and $C_t$ denote the covariance matrices of DL embeddings for DMSO-treated negative control wells in a source batch and the reference batch (Source~3), respectively.
The batch-corrected embeddings for all compounds in the source batch are obtained by applying the transformation:
\begin{equation}
  \hat{X}_s = X_s \, C_s^{-1/2} \, C_t^{1/2}
  \label{eq:coral}
\end{equation}
where $X_s$ is the matrix of original embeddings from the source batch.
By aligning on the DMSO controls that exhibit minimal compound-induced variation, the correction preserves biologically meaningful signals while removing technical variability.

\subsection{Molecule-Morphology Contrastive Pretraining}
\label{sec:contrastive}

Following the framework introduced in \mocop{} v1~\cite{Nguyen2023MoCoP}, we jointly learn a molecular encoder and a morphology encoder using contrastive learning on paired (molecule, morphology) data.
The pretraining dataset consists of $N$ molecule--morphology pairs $\{(x_i^{\text{mol}}, x_i^{\text{morph}}) \mid i \in \{1, \ldots, N\}\}$.

\textbf{Encoders and Projections.}
The molecular encoder $f_{\text{mol}}$ is a Gated Graph Neural Network (GGNN)~\cite{Li2016GGNN} that operates on 2D molecular graphs where atoms are nodes (with 75-dimensional features) and bonds are edges.
The GGNN consists of 6 message-passing layers followed by a fully connected layer of dimension 1024, with dropout ($p=0.1$).
The morphology encoder $f_{\text{morph}}$ is an MLP with hidden layer dimensions [512, 256, 128] and dropout ($p=0.1$), operating on the 128-dimensional batch-corrected DL embeddings.

Each encoder produces a representation that is transformed via a projection function $g$ into a shared embedding space:
\begin{align}
  h_i^{\text{mol}} &= f_{\text{mol}}(x_i^{\text{mol}}), \quad
  u_i^{\text{mol}} = \text{normalize}\bigl(g_{\text{mol}}(\sigma(h_i^{\text{mol}}))\bigr) \label{eq:enc_mol} \\
  h_i^{\text{morph}} &= f_{\text{morph}}(x_i^{\text{morph}}), \quad
  u_i^{\text{morph}} = \text{normalize}\bigl(g_{\text{morph}}(\sigma(h_i^{\text{morph}}))\bigr) \label{eq:enc_morph}
\end{align}
where $\sigma$ denotes a ReLU non-linearity, $g_{\text{mol}}$ and $g_{\text{morph}}$ are learned linear projections to $\mathbb{R}^{128}$, and $\text{normalize}(\cdot)$ denotes L2 normalization.

\textbf{Contrastive Objective.}
The two encoders are jointly optimized using a symmetric InfoNCE loss~\cite{Oord2018InfoNCE, Radford2021CLIP}.
Within a mini-batch of $N$ pairs, the cosine similarity matrix $S_{ij} = \langle u_i^{\text{mol}}, u_j^{\text{morph}} \rangle$ is computed, and the loss is defined as:
\begin{equation}
  \mathcal{L}_{\text{contrastive}} = \frac{1}{2}\bigl(\mathcal{L}_{\text{mol} \to \text{morph}} + \mathcal{L}_{\text{morph} \to \text{mol}}\bigr)
  \label{eq:symmetric_loss}
\end{equation}
where the directional losses are:
\begin{align}
  \mathcal{L}_{\text{mol} \to \text{morph}} &= -\frac{1}{N} \sum_{i=1}^{N} \log \frac{\exp(\tau \cdot S_{ii})}{\sum_{k=1}^{N} \exp(\tau \cdot S_{ik})} \label{eq:mol2morph} \\
  \mathcal{L}_{\text{morph} \to \text{mol}} &= -\frac{1}{N} \sum_{i=1}^{N} \log \frac{\exp(\tau \cdot S_{ii})}{\sum_{k=1}^{N} \exp(\tau \cdot S_{ki})} \label{eq:morph2mol}
\end{align}
with $\tau = 10$ as a temperature scaling parameter applied during training.

After pretraining, the molecular encoder $f_{\text{mol}}$ is transferred to downstream tasks, while the morphology encoder and projection heads are discarded.

\textbf{Training Details.}
The model was trained with a batch size of 256 using AdamW~\cite{Loshchilov2019AdamW} with cosine annealing warm restarts~\cite{Loshchilov2017SGDR}.
Early stopping was applied based on validation retrieval accuracy with a patience of 500 epochs.
The maximum training duration was set to 1,000 epochs.
Canonical SMILES representations were used for molecular identity resolution, and all training was conducted using PyTorch Lightning~\cite{Falcon2019PyTorchLightning}. \ref{app:mocop_hparams} describes additional details about training configurations.

\section{Results and Discussion}

\subsection{Morphology Retrieval Performance and Scaling Behavior}

We first evaluated the morphology retrieval performance of \mocop{} v2.
In this evaluation, each test compound's molecular graph is passed through the trained molecular encoder to produce a predicted embedding.
This predicted embedding is then compared against the morphological profiles of all compounds in a retrieval pool (a batch of either 100 or 1,000 test compounds), and the goal is to identify the correct morphological profile, i.e., the one belonging to the same compound.
Performance is measured by top-$k$ accuracy: the fraction of queries for which the correct match appears among the $k$ nearest neighbors ranked by cosine similarity in the shared embedding space.
Higher retrieval accuracy indicates that the molecular encoder has learned to map chemical structures to points in embedding space that are close to their corresponding morphological profiles.

To investigate how retrieval performance scales with training data, we trained \mocop{} v2 models using four subsets of the JUMP-CP dataset: 10K, 20K, 50K, and 98K (full) training compounds.
All models were trained from scratch with identical hyperparameters, using the same DL morphological embeddings and test split.

\begin{figure}[htbp]
  \centering
  \includegraphics[width=0.95\textwidth]{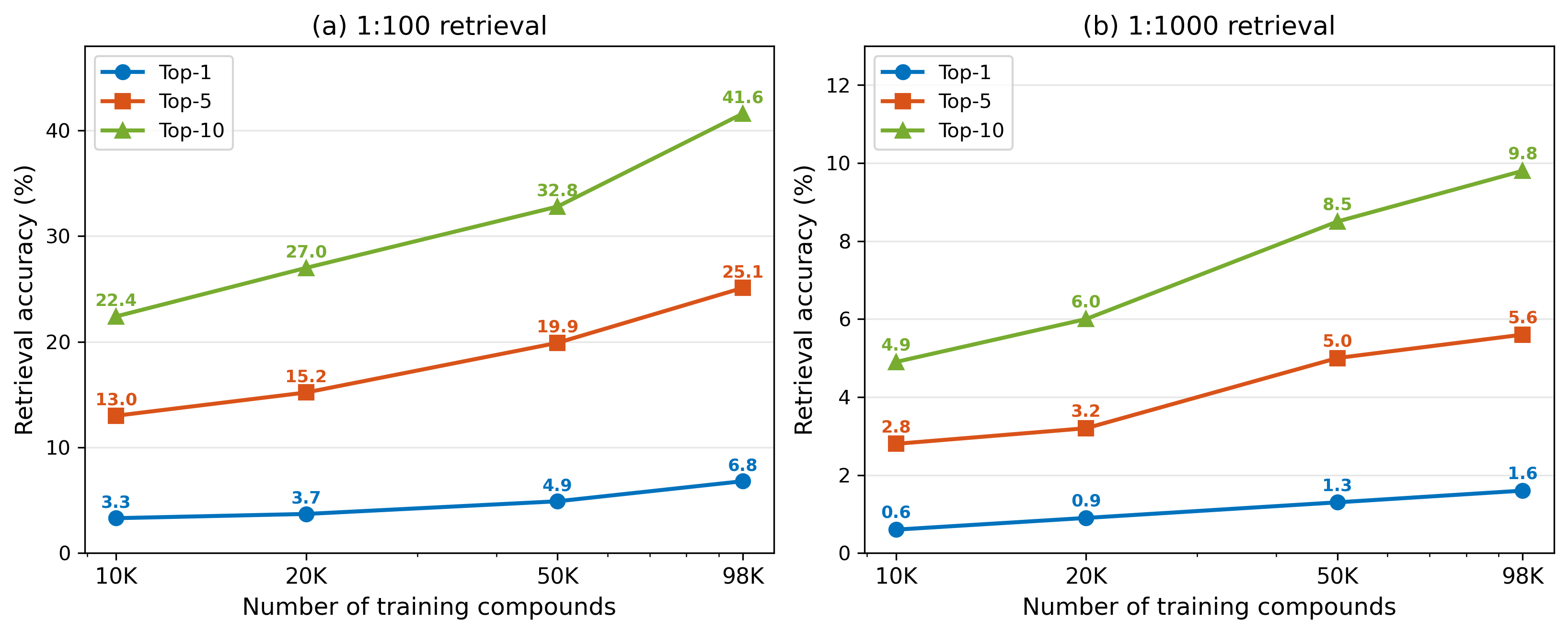}
  \caption{Scaling behavior of \mocop{} v2 morphology retrieval.
  Models are trained from scratch with 10K, 20K, 50K, and 98K compounds and evaluated on the same held-out test set.
  \textbf{(a)}~1:100 retrieval (100-candidate pool).
  \textbf{(b)}~1:1000 retrieval (1,000-candidate pool).
  Retrieval accuracy improves consistently with training data size across all top-$k$ metrics, following an approximately log-linear trend.}
  \label{fig:scaling}
\end{figure}

A clear scaling law is observed (Figure~\ref{fig:scaling}): retrieval accuracy improves consistently as more compounds are used for contrastive pretraining.
In the 1:100 retrieval setting, Top-1 accuracy increases from 3.3\% (10K compounds) to 6.8\% (98K compounds), a $2.1\times$ improvement.
Top-10 accuracy also rises from 22.4\% to 41.6\%, a $1.9\times$ improvement.
The 1:1000 retrieval setting exhibits a similar pattern, with Top-1 accuracy improving from 0.6\% to 1.6\% ($2.7\times$) and top-10 accuracy improving from 4.9\% to 9.8\% ($2\times$).
The approximately log-linear relationship between training data size and retrieval performance provides confidence that the framework can continue to benefit from additional data as larger Cell Painting datasets become available.

\subsection{MoCoP v2 Captures More Accurate Morphological Relationships}

Next, we compared the \mocop{} v1 and v2 embeddings in terms of how well they capture the morphological changes a compound introduces to the cells, and found \mocop{} v2 embeddings encode more biologically meaningful information about molecular perturbations than \mocop{} v1 embeddings through both qualitative and quantitative evaluations.
Pairwise cosine similarities between \mocop{} v1 and v2 embeddings are strongly correlated (Pearson $r = 0.873$, Spearman $\rho = 0.871$), indicating that the two models agree on most compound pairs.
However, the approximately 13\% of unexplained variance reveals cases where the models diverge, and in these cases, visual inspection of Cell Painting images consistently supports v2's assessment over v1's.

\textbf{Divergence Analysis: comparing raw Cell Painting images.}
To understand the differences between v1 and v2 embeddings, we identified compound pairs where the two models disagree most strongly about morphological similarity from the test dataset.
For each pair, we computed the cosine similarity in both the \mocop{} v1 and v2 embedding spaces after applying mean-centering to address similarity collapse, and selected pairs with the largest difference in similarity between the two models.
All image comparisons were restricted to Source~3 compounds not used for MoCoP training, eliminating confounding batch effects and keeping comparisons fair.

\begin{figure}[htbp]
  \centering
  \begin{subfigure}[t]{0.65\textwidth}
    \centering
    \includegraphics[width=\textwidth]{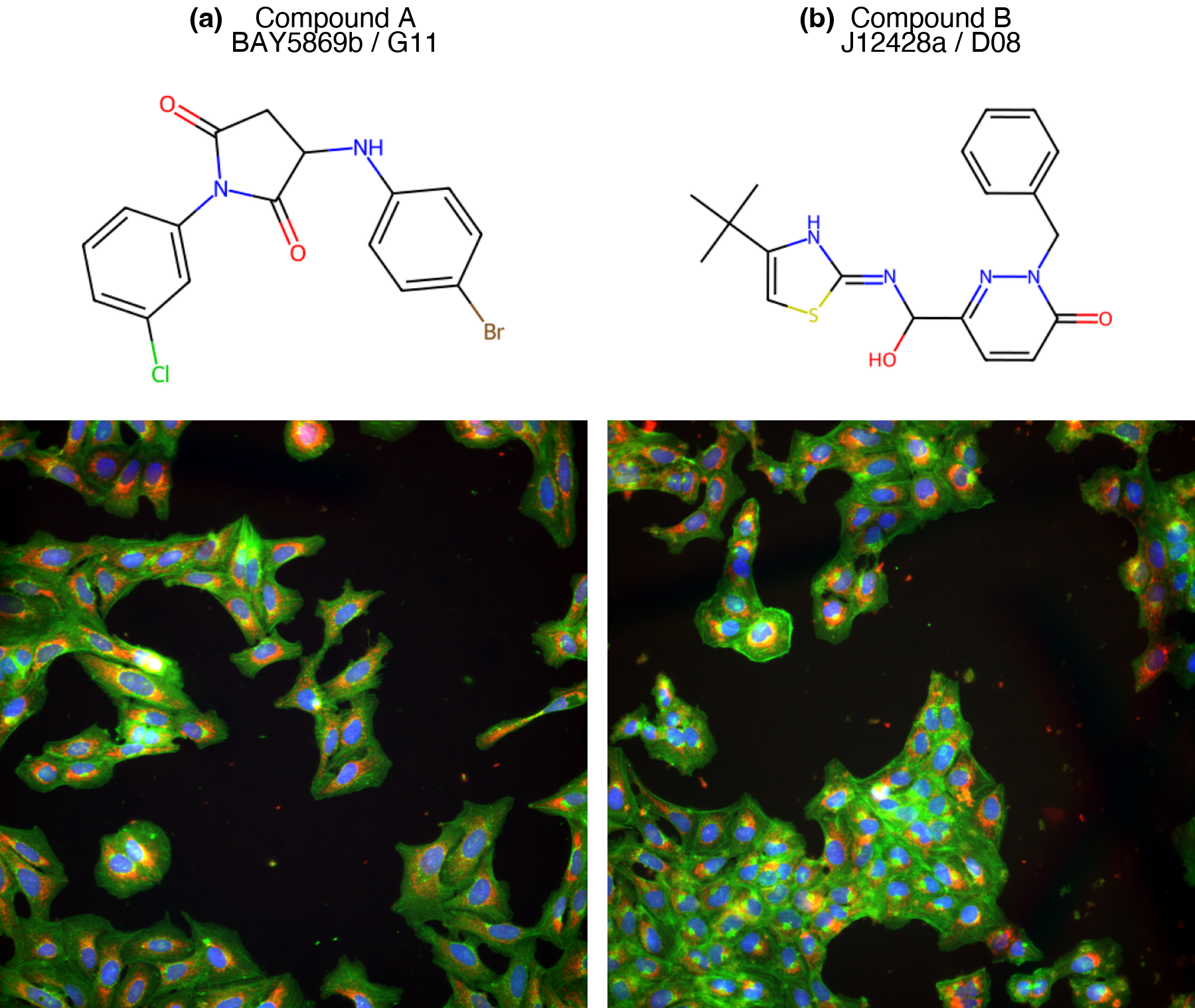}
    \caption{v1 incorrectly assigns high similarity to compounds with different phenotypes.
    Compound~A shows large, well-spread cells; Compound~B shows compact clusters with smaller cells
    (v1~cos~$= +0.619$, v2~cos~$= -0.387$, Tanimoto~$= 0.098$).}
  \end{subfigure}
  \\[1em]
  \begin{subfigure}[t]{0.65\textwidth}
    \centering
    \includegraphics[width=\textwidth]{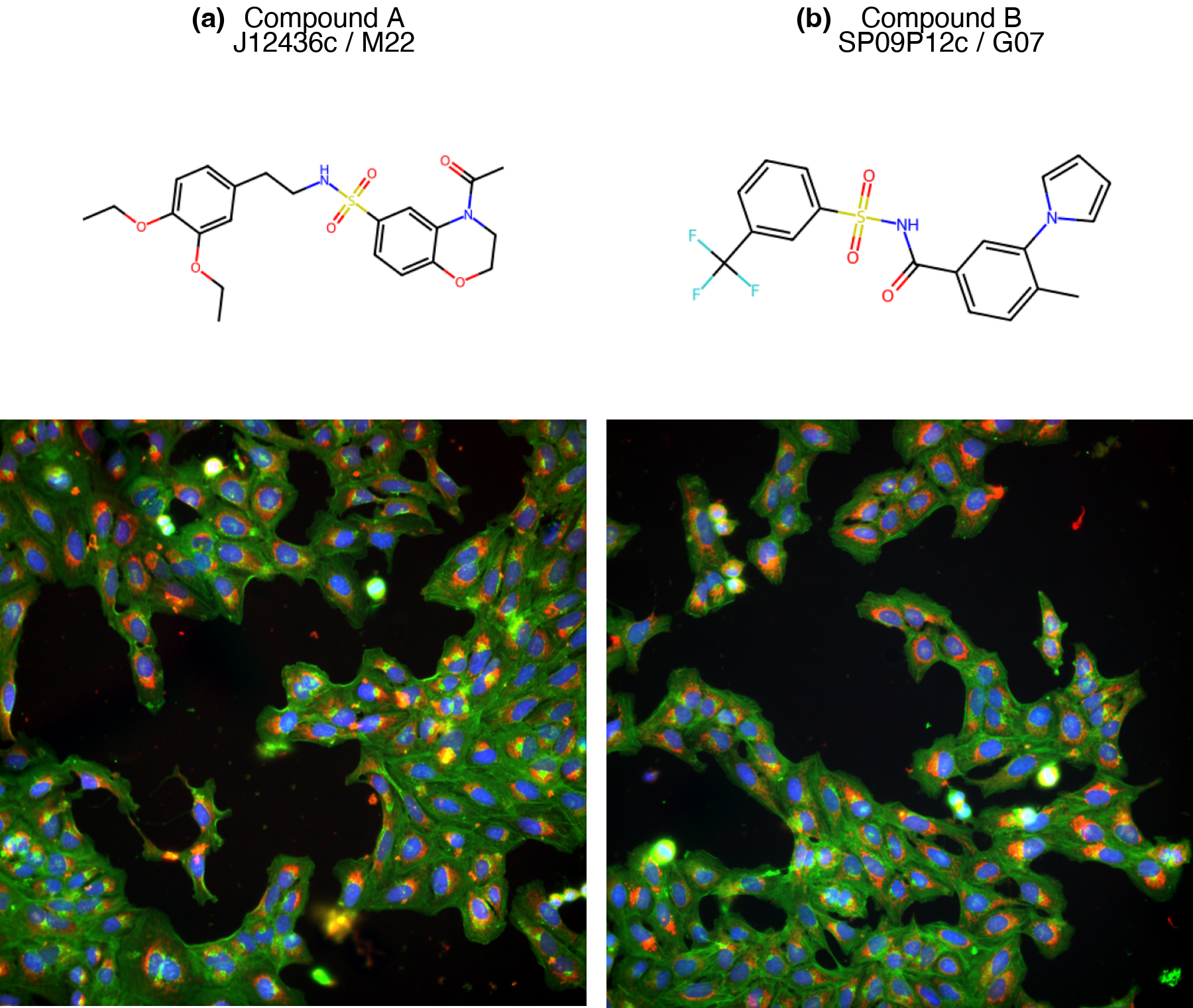}
    \caption{v2 correctly identifies phenotypic similarity missed by v1.
    Both compounds produce concordant phenotypes with similar cell density and channel balance
    (v2~cos~$= +0.615$, v1~cos~$= -0.286$, Tanimoto~$= 0.226$).}
  \end{subfigure}
  \caption{Representative examples of strong divergence between \mocop{} v1 and v2 embeddings.
  Cell Painting composite: Blue~=~DNA, Green~=~AGP, Red~=~Mitochondria.
  Chemical structures are shown alongside each compound. Plate barcodes and well positions from JUMP-CP are also marked for reference.
  \textbf{(a)}~v1 overestimates similarity for a pair with distinct phenotypes; v2 correctly identifies them as dissimilar.
  \textbf{(b)}~v2 recognizes shared phenotypic effects between structurally dissimilar compounds that v1 misses.}
  \label{fig:divergence}
\end{figure}

Figure~\ref{fig:divergence}a shows a representative case where \mocop{} v1 incorrectly assigns high similarity to two compounds that have visibly different phenotypes.
v1 assigns cosine similarity of $+0.619$ while v2 assigns $-0.387$.
The Cell Painting images reveal clear differences: Compound~A shows larger, well-spread cells at moderate density, while Compound~B produces compact cell clusters with smaller cell sizes and more visible disintegration.
v2 correctly identifies these as dissimilar.

Conversely, Figure~\ref{fig:divergence}b shows a case where \mocop{} v2 correctly identifies phenotypic similarity that v1 misses.
Despite a low Tanimoto similarity of 0.226, v2 assigns high similarity ($+0.615$) while v1 assigns low similarity ($-0.286$).
Inspection of the Cell Painting images confirms that both compounds produce concordant phenotypes (similar cell density, spreading pattern, and balance of contents from various visible channels), which is correctly captured by the DL-embedding-aligned v2 embeddings but missed by the CellProfiler-aligned v1 embeddings.

\ref{app:divergence_examples} shows additional examples of embedding divergence between \mocop{} v1 and v2, along with their corresponding Cell Painting images.
In all cases, the phenotypes observed in the images are more consistent with the v2 embedding similarities, confirming that v2 provides a more accurate representation of the morphological perturbations that compounds introduce to cells.

\textbf{Quantitative Context: Split-Half Replicate Concordance.}
To rigorously evaluate embedding quality, we designed a split-half replicate concordance analysis that measures how well each embedding's nearest neighbors capture \emph{reproducible} morphological similarity.
For each of the 5,265 test compounds with at least four replicate wells in the JUMP-CP dataset, we randomly split the wells into two independent halves (A and B) and computed compound-level CellProfiler profiles by averaging features within each half.
This yields two independent morphological similarity estimates for every compound pair.
We then established an \emph{oracle ceiling}: the retrieval quality achievable if molecular embeddings perfectly predicted morphological similarity.
Specifically, for each compound we selected its top-$k$ most morphologically similar neighbors using half~A profiles and evaluated those neighbors' morphological similarity using the independent half~B profiles.
This cross-validated ceiling represents the maximum retrieval quality that is reproducible across independent replicate measurements; it accounts for measurement noise and batch effects that limit any model's achievable performance.

\begin{figure}[htbp]
  \centering
  \begin{subfigure}[t]{0.48\textwidth}
    \centering
    \includegraphics[width=\textwidth]{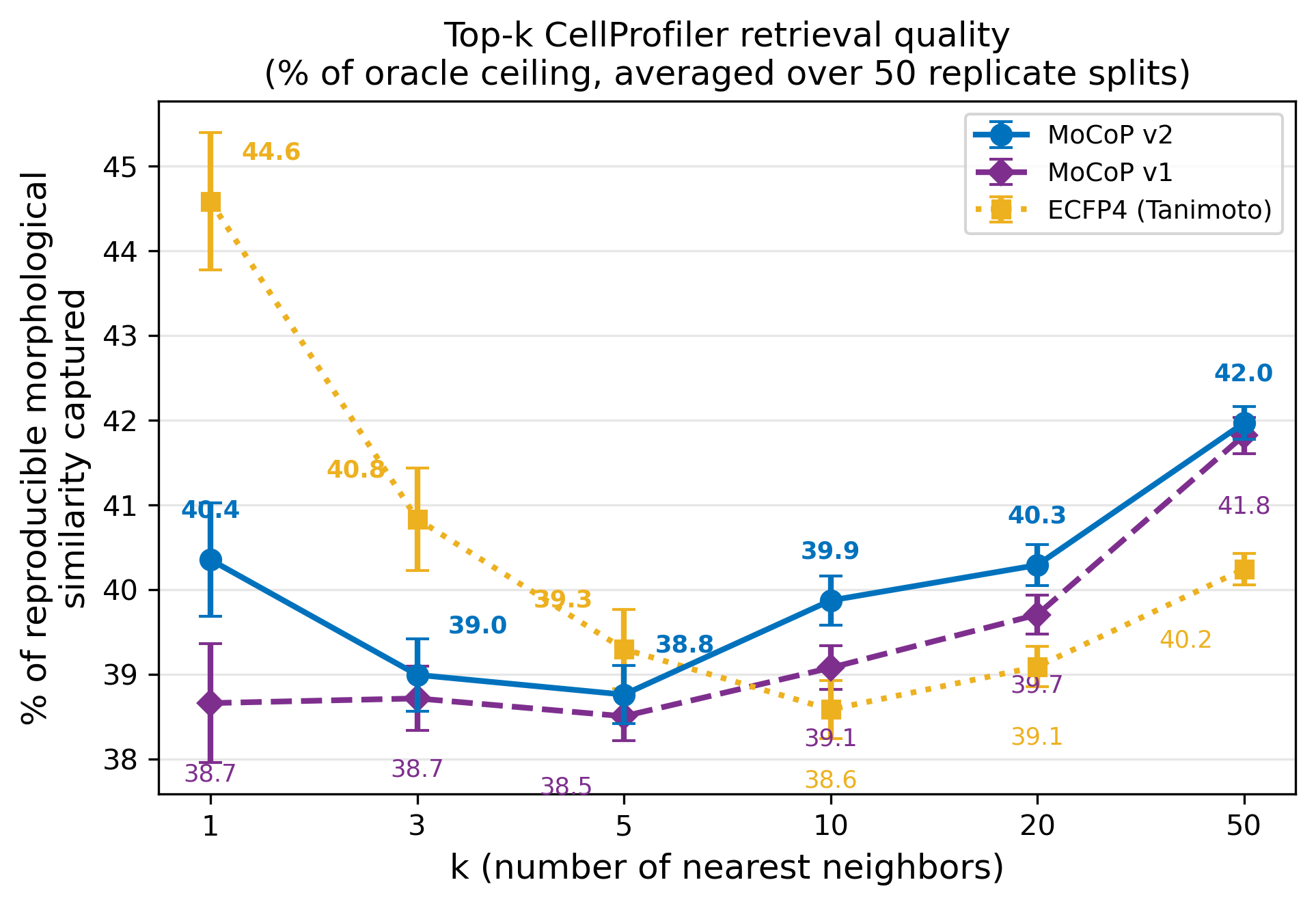}
    \caption{Top-$k$ CellProfiler retrieval (\% of oracle ceiling)}
  \end{subfigure}
  \hfill
  \begin{subfigure}[t]{0.48\textwidth}
    \centering
    \includegraphics[width=\textwidth]{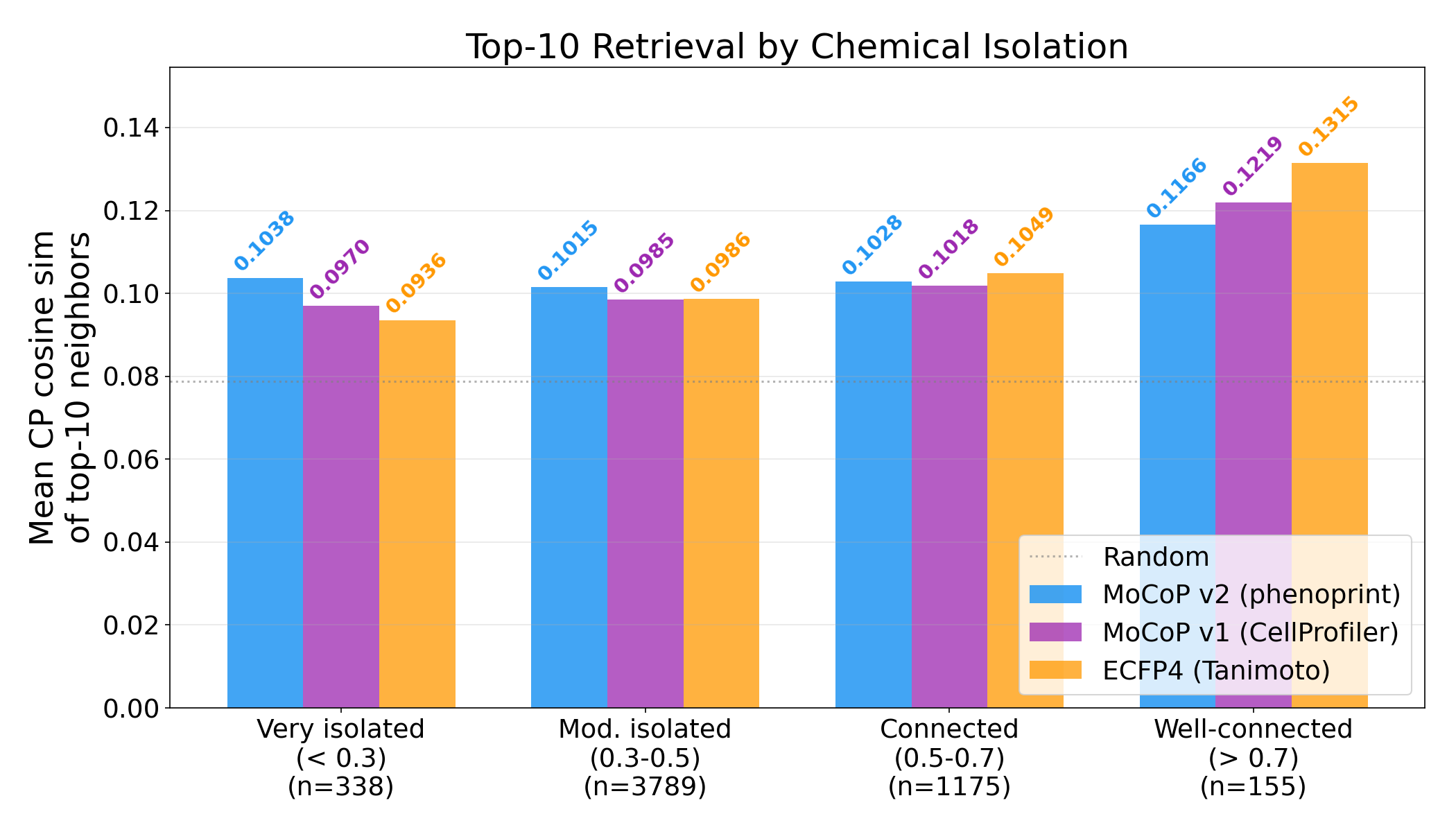}
    \caption{Stratified by chemical isolation ($k=10$)}
  \end{subfigure}
  \caption{Quantitative comparison of \mocop{} v2, v1, and ECFP4 embedding quality, measured by CellProfiler profile retrieval on 5,265 JUMP test compounds with $\geq 4$ replicate wells.
  \textbf{(a)}~Retrieval quality expressed as percentage of the oracle ceiling (see text), averaged over 50 random replicate splits.
  Error bars indicate bootstrap standard deviation.
  ECFP4 dominates at small $k$ but declines as structural analogs are exhausted; v2 overtakes ECFP4 at $k \geq 10$ and consistently outperforms v1 at all $k$ values.
  \textbf{(b)}~v2 provides the best morphology retrieval for chemically isolated compounds (Tanimoto $< 0.3$), exactly the regime where structure-based methods have the least information.}
  \label{fig:quantitative_context}
\end{figure}

For each embedding method (\mocop{} v2, \mocop{} v1, and ECFP4 with 1024-bit Morgan fingerprints), we retrieved the top-$k$ nearest neighbors in embedding space and measured the mean CellProfiler cosine similarity of those neighbors, averaged over both replicate halves.
Results are expressed as a percentage of the oracle ceiling to normalize for the inherent difficulty of the task at each $k$ (Figure~\ref{fig:quantitative_context}a).
The entire procedure was repeated across 50 random replicate splits to obtain bootstrap confidence intervals; the resulting error bars are small, confirming that the observed differences are robust to the choice of replicate partition.
Neither \mocop{} v1 nor v2 was trained on the CellProfiler features used for evaluation, making this an independent assessment of embedding quality.

ECFP4 achieves the highest retrieval quality at small $k$ ($44.6\%$ of ceiling at $k=1$), reflecting that structurally similar compounds often induce similar phenotypes.
However, ECFP4's performance declines as $k$ increases ($38.6\%$ at $k=10$) because it exhausts the pool of close structural analogs.
\mocop{} v2 maintains more stable performance across all $k$ values ($38.8$--$42.0\%$ of ceiling) and overtakes ECFP4 at $k \geq 10$, demonstrating its ability to identify phenotypically similar compounds beyond the reach of structural fingerprints.
\mocop{} v2 also consistently outperforms v1 at every $k$ value tested, with the strongest advantage at $k=1$, showing the v2 embeddings are more capable of accurately identifying phenotypically similar compounds with immediate neighbors in the embedding space.
This is especially notable given that \mocop{} v1 was explicitly aligned to CellProfiler profiles, yet v2, which was aligned to deep-learning morphological embeddings, better predicts CellProfiler-based similarity.

The advantage of v2 is most profound for \emph{chemically isolated} compounds (maximum Tanimoto similarity to any other compound $< 0.3$).
In this regime, where structural fingerprints provide limited information, MoCoP v2 outperforms both MoCoP v1 and ECFP4 (Figure~\ref{fig:quantitative_context}b) significantly.
This is precisely where morphology-informed embeddings provide the greatest value: identifying molecules with distinct scaffolds that nonetheless share similar phenotypic effects.
Indeed, even among compound pairs with Tanimoto similarity below 0.2, \mocop{} v2 can still identify shared or divergent phenotypic effects that v1 misses.
\ref{app:divergence_examples} presents visual examples of such structurally dissimilar pairs where v2 captures morphological relationships invisible to both v1 and chemical fingerprints, highlighting the value of DL-embedding-aligned molecular representations for scaffold hopping applications.

\subsection{Improved QSAR Prediction on ChEMBL20}

We evaluated the transferability of \mocop{} v2 molecular embeddings on QSAR prediction tasks using the ChEMBL20 dataset\cite{Mayr2018Chembl}, following the exact procedure as described in MoCoP v1. ChEMBL20 dataset comprises a diverse set of measurements including ADME, toxicity, physicalchemical properties, binding and functional assays, consisting of 1,310 binary downstream tasks of about 450K compounds.
Two evaluation protocols were used for comparison: (1)~\emph{linear probing}, where the pretrained molecular encoder is frozen and a single linear layer is trained on top; and (2)~\emph{end-to-end fine-tuning}, where the full molecular encoder is updated jointly with a task-specific head, with molecular encoder weights initialized from the pretrained checkpoint.
Both protocols were evaluated across three data regimes (5\%, 25\%, and 100\% of the ChEMBL20 data) to assess data efficiency.
Results are reported as mean $\pm$ standard deviation across 3 train/validation/test splits.

\textbf{Linear Probe Results.}
Table~\ref{tab:linear_probe} shows that \mocop{} v2 consistently outperforms \mocop{} v1 across all data proportions for both AUROC and AUPRC, with an improvement of 3.5 percentage points in AUROC and 2.4 in AUPRC at 5\% data; 3.1 and 2.9 at 25\% data; and 2.5 and 2.6 at 100\% data.
These consistent gains confirm that the DL-embedding-aligned representations from \mocop{} v2 encode more informative molecular features than CellProfiler-aligned embeddings, which could be easily exploited by a linear model from the frozen embeddings.

\begin{table}[htbp]
  \caption{Linear probe QSAR prediction performance on ChEMBL20. Mean $\pm$ standard deviation across 3 splits. Bold indicates the better result.}
  \label{tab:linear_probe}
  \centering
  \begin{tabular}{l cccc}
    \toprule
    Data & \multicolumn{2}{c}{AUROC} & \multicolumn{2}{c}{AUPRC} \\
    \cmidrule(lr){2-3} \cmidrule(lr){4-5}
    Proportion & \mocop{} v1 & \mocop{} v2 & \mocop{} v1 & \mocop{} v2 \\
    \midrule
    5\%   & $0.570 \pm 0.025$ & $\mathbf{0.605 \pm 0.022}$ & $0.552 \pm 0.036$ & $\mathbf{0.576 \pm 0.028}$ \\
    25\%  & $0.637 \pm 0.009$ & $\mathbf{0.668 \pm 0.032}$ & $0.602 \pm 0.027$ & $\mathbf{0.631 \pm 0.037}$ \\
    100\% & $0.706 \pm 0.041$ & $\mathbf{0.731 \pm 0.026}$ & $0.669 \pm 0.044$ & $\mathbf{0.695 \pm 0.039}$ \\
    \bottomrule
  \end{tabular}
\end{table}

\textbf{End-to-End Fine-Tuning Results.}
When allowing the full molecular encoder to be fine-tuned on the downstream QSAR tasks, the results become more nuanced (Table~\ref{tab:finetuning}).
\mocop{} v2 slightly underperforms v1 in the low-data regimes (5\% and 25\% AUROC) but surpasses v1 at 100\% data, with improvements of 1.3 percentage points in AUROC and 3.2 in AUPRC.
v2 achieves higher AUPRC than v1 across all data proportions, indicating better precision--recall trade-offs.
Even though there are some overlap in the confidence intervals, the consistent direction of improvement in AUPRC across all three data proportions, combined with the similar gains observed in the linear probe setting (Table~\ref{tab:linear_probe}), supports the conclusion that v2 embeddings encode richer information.
The fine-tuning results indicate that this richer information is most effectively exploited when sufficient downstream data is available for adaptation.

\begin{table}[htbp]
  \caption{End-to-end fine-tuning QSAR prediction performance on ChEMBL20. Mean $\pm$ standard deviation across 3 splits. Bold indicates the better result.}
  \label{tab:finetuning}
  \centering
  \begin{tabular}{l cccc}
    \toprule
    Data & \multicolumn{2}{c}{AUROC} & \multicolumn{2}{c}{AUPRC} \\
    \cmidrule(lr){2-3} \cmidrule(lr){4-5}
    Proportion & \mocop{} v1 & \mocop{} v2 & \mocop{} v1 & \mocop{} v2 \\
    \midrule
    5\%   & $\mathbf{0.621 \pm 0.022}$ & $0.598 \pm 0.015$ & $0.569 \pm 0.023$ & $\mathbf{0.575 \pm 0.025}$ \\
    25\%  & $\mathbf{0.689 \pm 0.018}$ & $0.677 \pm 0.013$ & $0.640 \pm 0.031$ & $\mathbf{0.648 \pm 0.024}$ \\
    100\% & $0.721 \pm 0.020$ & $\mathbf{0.734 \pm 0.034}$ & $0.681 \pm 0.033$ & $\mathbf{0.713 \pm 0.040}$ \\
    \bottomrule
  \end{tabular}
\end{table}

\subsection{Comparing with Other State-of-the-Art Methods}

To contextualize the performance of \mocop{} v2 against existing state-of-the-art molecular embedding methods that may incorporate one or more modalities, including molecular structure, cell morphology and transcriptomics, we evaluated on four benchmark datasets spanning diverse prediction tasks: ToxCast (toxicity classification: 8,576 compounds across 617 tasks)\cite{toxcast}, ChEMBL~2K (activity classification: 2,355 compounds across 41 tasks)\cite{chembl2k}, Broad~6K (activity classification: 6,567 compounds across 32 tasks)\cite{Becker2020PredictingActivity}, and Biogen~3K (ADME regression: 3,521 compounds across 6 tasks)\cite{biogen3k}.
These benchmarks were curated from the InfoAlign repository~\cite{Liu2024InfoAlign} and are widely used for evaluating various molecular representations. Train, validation and test splits also follow the same scaffold-based splitting as in~\cite{Liu2024InfoAlign}, which allows fair comparison across methods.
All results are reported as 3-seed mean $\pm$ standard deviation on the test set.
A comprehensive hyperparameter sweep was conducted for \mocop{} v2 to identify optimal configurations for each dataset (see~\ref{app:downstream_details}).

Table~\ref{tab:benchmark} compares \mocop{} v2 with several baseline and state-of-the-art methods.
To facilitate interpretation, the table is organized into three groups based on the training data modalities.
The first group consists of \emph{structure-only} baselines (an MLP trained on Morgan fingerprints~\cite{Rogers2010ECFP}, MolT5~\cite{Edwards2022MolT5}, and Uni-Mol~\cite{Zhou2023UniMol}), which use only molecular structure during training.
The second group contains methods that use \emph{structure and cell morphology only}: CLOOME~\cite{SanchezFernandez2023CLOOME}, which aligns molecular graphs with Cell Painting images, and \mocop{} v2.
The third group includes InfoAlign~\cite{Liu2024InfoAlign} and CHMR~\cite{Li2025CHMR}, which use \emph{structure, cell morphology, and transcriptomics} during training.
This distinction is important for a fair comparison: InfoAlign constructs a context graph that jointly encodes molecular, morphological, and gene expression modalities, while CHMR models hierarchical dependencies across molecular, cellular, and genomic levels using vector quantization.
Both methods therefore have access to additional biological information (transcriptomic readouts) that is unavailable to \mocop{} v2 during training.

\begin{table}[htbp]
  \caption{Performance comparison on downstream benchmark tasks.
  Classification tasks (ChEMBL~2K, ToxCast, Broad~6K) report average AUC~(\%, $\uparrow$).
  Biogen~3K reports average MAE~($\times 100$, $\downarrow$).
  \textbf{Bold} indicates the best result; \underline{underline} indicates the second best.
  Methods are grouped by training data modalities.
  Values for baselines, CLOOME, InfoAlign, and CHMR are taken from~\cite{Liu2024InfoAlign, Li2025CHMR}.}
  \label{tab:benchmark}
  \centering
  \begin{tabular}{lcccc}
    \toprule
    Method & ChEMBL~2K AUC $\uparrow$ & ToxCast AUC $\uparrow$ & Broad~6K AUC $\uparrow$ & Biogen~3K MAE $\downarrow$ \\
    \midrule
    \multicolumn{5}{l}{\textit{Structure only}} \\
    MLP & $76.8 \pm 2.2$ & $57.6 \pm 1.0$ & $63.3 \pm 0.3$ & $66.2 \pm 2.4$ \\
    MolT5\cite{Edwards2022MolT5} & $69.9 \pm 0.8$ & $64.7 \pm 0.9$ & $55.1 \pm 0.9$ & $65.1 \pm 0.5$ \\
    UniMol\cite{Zhou2023UniMol} & $76.8 \pm 0.4$ & $64.6 \pm 0.2$ & $65.4 \pm 0.1$ & $55.8 \pm 2.8$ \\
    \midrule
    \multicolumn{5}{l}{\textit{Structure + cell morphology}} \\
    CLOOME\cite{SanchezFernandez2023CLOOME} & $66.7 \pm 1.8$ & $54.2 \pm 0.9$ & $61.7 \pm 0.4$ & $64.3 \pm 0.4$ \\
    \mocop{} v2 & $\underline{81.5 \pm 0.6}$ & $\mathbf{70.3 \pm 0.2}$ & $68.9 \pm 1.2$ & $\underline{43.5 \pm 0.4}$ \\
    \midrule
    \multicolumn{5}{l}{\textit{Structure + cell morphology + transcriptomics}} \\
    InfoAlign\cite{Liu2024InfoAlign} & $81.3 \pm 0.6$ & $66.4 \pm 1.1$ & $\underline{70.0 \pm 0.1}$ & $49.4 \pm 0.2$ \\
    CHMR\cite{Li2025CHMR} & $\mathbf{84.7 \pm 0.2}$ & $\underline{69.3 \pm 0.3}$ & $\mathbf{71.4 \pm 0.2}$ & $\mathbf{40.9 \pm 0.3}$ \\
    \bottomrule
  \end{tabular}
\end{table}

\mocop{} v2 achieves the \textbf{best performance on ToxCast} ($70.3 \pm 0.2$), outperforming CHMR ($69.3 \pm 0.3$) by 1.0 percentage point and InfoAlign ($66.4 \pm 1.1$) by 3.9 percentage points. Notably, \mocop{} v2 achieves this using only cell morphology, whereas both CHMR and InfoAlign additionally leverage transcriptomic data during training.
On Biogen~3K, \mocop{} v2 ($43.5 \pm 0.4$ MAE) ranks second after CHMR ($40.9 \pm 0.3$) and substantially outperforms InfoAlign ($49.4 \pm 0.2$).
On ChEMBL~2K, \mocop{} v2 ($81.5 \pm 0.6$) matches InfoAlign and trails CHMR by 3.2 points.
On Broad~6K, \mocop{} v2 ($68.9 \pm 1.2$) slightly trails both InfoAlign and CHMR, but still outperforms all structure-only baselines and CLOOME.

Overall, \mocop{} v2 outperforms InfoAlign on 3 out of 4 benchmarks (ToxCast, ChEMBL~2K, and Biogen~3K) and achieves the best result on ToxCast across all methods.
These results are particularly noteworthy given the architectural simplicity of \mocop{} (a GGNN molecular encoder with contrastive alignment to DL morphological embeddings), compared with the multi-modal context graphs of InfoAlign or the hierarchical vector quantization of CHMR.
Where CHMR outperforms \mocop{} v2 (ChEMBL~2K, Broad~6K, and most Biogen~3K tasks), the additional transcriptomic signal and more expressive architecture likely contribute to its advantage. However, \mocop{} v2's competitive performance using morphology alone suggests that cell morphology captures a substantial portion of the biological information relevant to these tasks.

\textbf{Biogen~3K Per-Task Analysis.}
A per-task breakdown of the Biogen~3K ADME regression results reveals complementary strengths (Table~\ref{tab:biogen_pertask}).
\mocop{} v2 achieves the best performance on 2 of 6 tasks: MDR1-MDCK efflux ratio ($33.8 \pm 0.3$) and rat liver microsomal clearance ($38.8 \pm 0.4$), which again show the strength of the method for certain ADME predictions compared with a much more complicated architecture trained with additional transcriptomics information.
Notably, \mocop{} v2 also outperforms InfoAlign on all 6 individual tasks.

\begin{table}[htbp]
  \caption{Biogen~3K per-task comparison (MAE $\times 100$, $\downarrow$). Bold indicates the best result.}
  \label{tab:biogen_pertask}
  \centering
  \begin{tabular}{lccc}
    \toprule
    Task & \mocop{} v2 & CHMR\cite{Li2025CHMR} & InfoAlign\cite{Liu2024InfoAlign} \\
    \midrule
    LOG HLM CLint & $35.4 \pm 1.0$ & $\mathbf{33.7 \pm 0.4}$ & $39.7 \pm 0.4$ \\
    LOG MDR1-MDCK ER & $\mathbf{33.8 \pm 0.3}$ & $35.2 \pm 0.2$ & $39.2 \pm 0.3$ \\
    LOG Solubility @ pH 6.8 & $37.4 \pm 0.5$ & $\mathbf{34.9 \pm 0.5}$ & $40.5 \pm 0.6$ \\
    LOG PPB Human & $60.6 \pm 2.1$ & $\mathbf{53.1 \pm 1.3}$ & $66.7 \pm 1.7$ \\
    LOG PPB Rat & $55.1 \pm 1.1$ & $\mathbf{48.5 \pm 0.9}$ & $62.0 \pm 1.5$ \\
    LOG RLM CLint & $\mathbf{38.8 \pm 0.4}$ & $39.8 \pm 0.3$ & $48.4 \pm 0.6$ \\
    \midrule
    Average & $43.5 \pm 0.4$ & $\mathbf{40.9 \pm 0.3}$ & $49.4 \pm 0.2$ \\
    \bottomrule
  \end{tabular}
\end{table}

\section{Conclusion}

We have presented \mocop{} v2, an improved molecular-morphology contrastive pretraining framework that replaces CellProfiler-derived morphological features with deep-learning-based morphological embeddings as alignment targets.
The DL embedding pipeline, comprising a fine-tuned ResNet-18 encoder and CORAL batch correction, extracts richer morphological representations from Cell Painting images, leading to improved molecular embeddings after alignment.

Our key findings are: (1)~\mocop{} v2 molecular embeddings more accurately capture how molecules perturb cell morphology, even outperforming CellProfiler-aligned embeddings on CellProfiler-based similarity retrieval; (2)~the pretrained embeddings consistently improve QSAR prediction in linear probe evaluations, with gains across all data proportions; (3)~\mocop{} v2 achieves state-of-the-art performance on ToxCast toxicity prediction and competitive results on ADME and activity benchmarks compared to recent methods that additionally use transcriptomic data; and (4)~retrieval performance follows a log-linear scaling law with training data size, with consistent improvements from 10K to 98K training compounds.

The pretrained molecular encoder from \mocop{} v2 serves as a readily tunable foundation for a broad range of \emph{in silico} drug design applications.
As demonstrated in this work, the encoder can be efficiently adapted via either linear probing or end-to-end fine-tuning to QSAR and ADMET prediction tasks that are central to lead optimization and safety assessment in drug discovery pipelines.
Because the encoder has internalized morphology-grounded information about how molecules perturb living cells, it provides a complementary signal to purely structure-derived descriptors, particularly for chemically novel scaffolds where structural fingerprints offer limited predictive power.
Beyond property prediction, the learned embeddings are also applicable to mechanism-of-action (MOA) retrieval: by querying nearest neighbors in the \mocop{} embedding space, one can identify compounds that induce similar cellular phenotypes regardless of structural similarity, supporting scaffold hopping and target deconvolution efforts.

Looking ahead, several avenues for further improvement become feasible.
On the DL embedding extraction side, the ResNet-18 backbone could be replaced by self-supervised vision transformers such as DINO~\cite{Doron2023DINO, CrossZamirski2022WSDINO}, which have demonstrated superior performance on Cell Painting data by learning attention-based representations that capture structurally meaningful phenotypic features at both subcellular and population scales without requiring manual annotations.
Masked autoencoders~\cite{Kraus2024MAE, Kraus2023MAE} offer another promising direction, having shown scalable improvements with larger models and datasets on microscopy images.
On the contrastive pretraining side, the current architecture treats the molecule and morphology modalities as separate encoders bridged by a cosine-similarity objective.
An alternative approach would be to model the joint distribution of molecular tokens and morphological features through a unified transformer architecture with masked token prediction, analogous to how masked language models learn bidirectional context, which could capture richer cross-modal dependencies than the current contrastive alignment.
The molecular encoder itself could also be upgraded from a GGNN to a more expressive architecture, such as a graph transformer.
Multi-modal extensions incorporating transcriptomic data~\cite{Way2022Complementary, Ha2025CrossModality} could provide additional biological context, as gene expression and morphology capture complementary aspects of cellular state.
Additionally, the current framework does not account for compound concentration: the same molecule tested at different doses can produce markedly different phenotypic responses, yet \mocop{} v2 maps each compound to a single embedding regardless of dose.
Incorporating concentration-aware embeddings, for example by conditioning the molecular encoder on the treatment dose, could enable the model to capture dose--response relationships and improve predictions for concentration-dependent endpoints.
Similarly, cell-line-specific or tissue-specific fine-tuning of the morphological encoder could tailor the learned representations to particular biological contexts, potentially improving performance on tissue-specific assays.
Finally, the growing availability of large-scale Cell Painting datasets from the JUMP consortium and other sources~\cite{Chandrasekaran2023JUMPCP, Fay2023RxRx3} offers opportunities for continued scaling, which our results suggest will yield further improvements in representation quality.

\section*{Acknowledgements}

The authors thank Yu Yan and Lawrence Du from GSK AIML for their inspiring discussions and valuable feedbacks. The authors also would like to thank the JUMP Cell Painting Consortium for making large-scale Cell Painting data publicly available, and the GSK Onyx platform for support on the computation infrastructure. This manuscript has used Claude Opus 4.6 model for language editing and formatting assistance.

\section*{Data and Code Availability}

The JUMP-CP Cell Painting data are publicly available through the Cell Painting Gallery (\url{https://registry.opendata.aws/cellpainting-gallery}).
The downstream benchmark datasets (ChEMBL~2K, ToxCast, Broad~6K, Biogen~3K) were obtained from the InfoAlign repository (\url{https://github.com/liugangcode/InfoAlign}). The DL morphological embedding pipeline is a proprietary cell imaging processing workflow, and the training of MoCoP v2 reuses the codebase for MoCoP v1, which is available at \url{https://github.com/GSK-AI/mocop}.

\printbibliography

\clearpage
\appendix
\renewcommand{\thesection}{Appendix \Alph{section}}
\setcounter{secnumdepth}{1}
\setcounter{section}{0}

\section{: DL Embedding Training Hyperparameters}
\label{app:dl_emb_hparams}

Table~\ref{tab:dl_emb_hparams} summarizes the key hyperparameters used for training the DL morphological embedding image encoder.

\begin{table}[htbp]
  \caption{DL morphological embedding image encoder training hyperparameters.}
  \label{tab:dl_emb_hparams}
  \centering
  \begin{tabular}{ll}
    \toprule
    Hyperparameter & Value \\
    \midrule
    Backbone & ResNet-18 (ImageNet pretrained) \\
    Projection head & MLP: $\text{backbone} \to 1024 \to 128$ \\
    Embedding dimension & 128 \\
    Tiles per well & 12 \\
    Aggregation & Mean \\
    Loss function & Triplet margin loss (margin = 0.1) \\
    Mining strategy & Semihard \\
    Classes per batch & 8 \\
    Oversampling multiplier & 2 \\
    Batch size & 64 \\
    Optimizer & AdamW \\
    Learning rate & $1 \times 10^{-5}$ \\
    Weight decay & $2 \times 10^{-3}$ \\
    Epochs & 100 \\
    Training data & JUMP-CP Source~3 (217 plates, 54,655 wells) \\
    Evaluation metric & 1-NN Euclidean accuracy \\
    \bottomrule
  \end{tabular}
\end{table}

\subsubsection{Data Augmentation.}
Training images were augmented with random horizontal/vertical flips ($p=0.5$), random 90-degree rotations ($p=0.5$), per-channel multiplicative noise (factors uniformly sampled from $[0.9, 1.1]$), and per-channel normalization.
Validation and test images received only per-channel normalization.

\clearpage
\section{: Dataset Summary Statistics}
\label{app:dataset_stats}

Table~\ref{tab:dataset_stats} summarizes the datasets used for DL embedding extraction, contrastive pretraining, and evaluation.
The JUMP-CP consortium data spans 10 sources (independent laboratories), each contributing varying numbers of plates and compounds.
Compound-based splitting ensures that no compound appears in both the training and test sets; the split is applied with the canonical SMILES to avoid data leakage from alternative SMILES representations.

\begin{table}[htbp]
  \caption{Summary of datasets used in the \mocop{} v2 pipeline.
  ``Wells'' refers to the number of compound-treated wells (excluding DMSO controls used only for CORAL correction).
  ``Compounds'' counts unique chemical entities (by canonical SMILES).}
  \label{tab:dataset_stats}
  \centering
  \begin{tabular}{llrrr}
    \toprule
    Stage & Dataset & Wells & Compounds & Plates \\
    \midrule
    \multicolumn{5}{l}{\textit{DL image encoder training}} \\
    & JUMP-CP Source~3 & 66,471 & 55,336 & 217 \\
    \midrule
    \multicolumn{5}{l}{\textit{Contrastive pretraining (DL embeddings)}} \\
    & Source~3 only & 66,471 & 55,336 & 217 \\
    & Full JUMP-CP (10 sources) & 649,384 & 109,164 & 1,685 \\
    \midrule
    \multicolumn{5}{l}{\textit{Compound splits (split 0)}} \\
    & Source~3 train & & 49,785 & \\
    & Source~3 validation & & 2,717 & \\
    & Source~3 test & & 2,809 & \\
    & Full JUMP train & & 98,247 & \\
    & Full JUMP validation & & 5,459 & \\
    & Full JUMP test & & 5,459 & \\
    \bottomrule
  \end{tabular}
\end{table}

\clearpage
\section{: CORAL Batch Correction Visualization}
\label{app:coral}

\begin{figure}[htbp]
  \centering
  \includegraphics[width=0.95\textwidth]{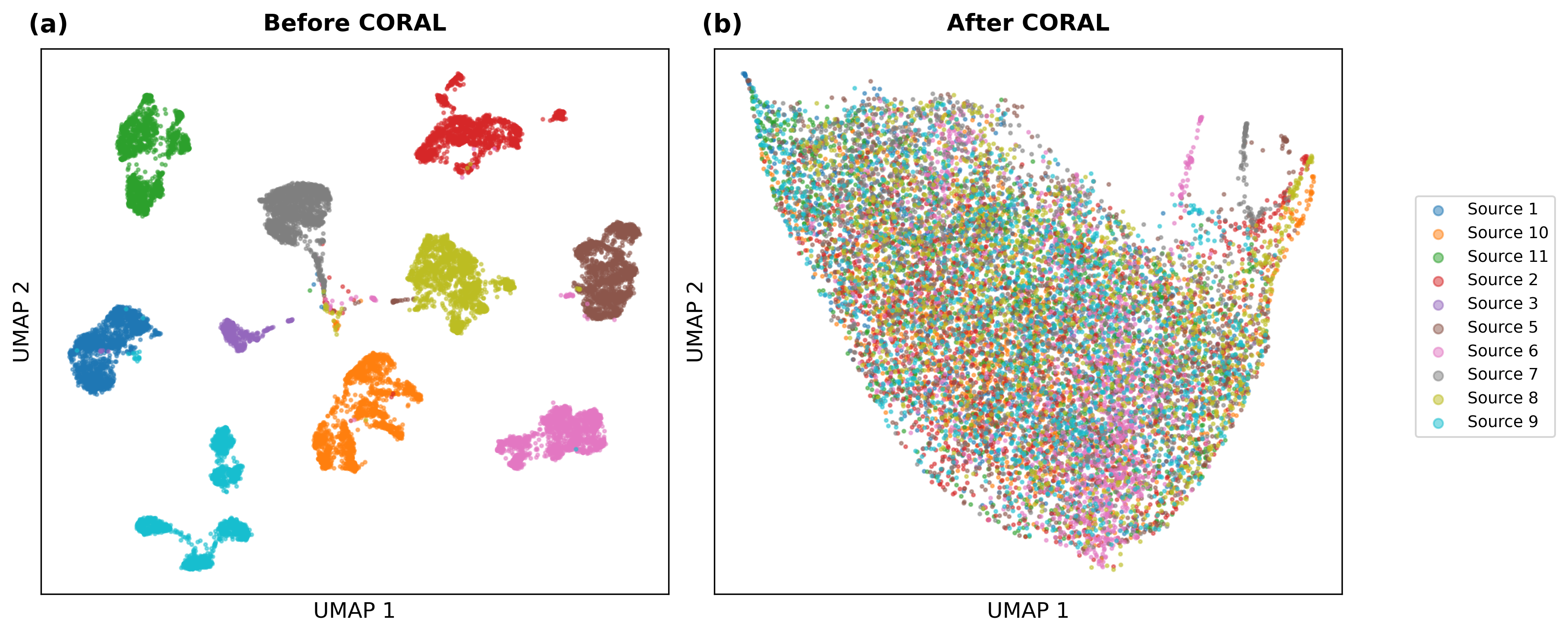}
  \caption{Effect of CORAL batch correction on DL morphological embeddings.
  UMAP projections (cosine distance, $n_{\text{neighbors}}=30$) colored by JUMP-CP data source, subsampled to $\sim$1,500 wells per source.
  \textbf{(a)}~Before CORAL correction: embeddings cluster by data source. Batch effects dominate the clustering.
  \textbf{(b)}~After CORAL correction using DMSO negative controls: source-specific clusters dissolve, and embeddings are intermixed across sources.
  Samples with outlier embedding norms ($<$1\% of data) were excluded from the corrected panel.}
  \label{fig:coral_before_after}
\end{figure}

To illustrate the necessity and effectiveness of CORAL batch correction, Figure~\ref{fig:coral_before_after} shows UMAP projections of DL embeddings colored by data source, before and after CORAL correction.
Before correction, the embeddings cluster strongly by data source, indicating that batch effects dominate the representation.
After CORAL correction using DMSO negative controls, the source-specific clusters dissolve, as can be seen in Figure~\ref{fig:coral_before_after}(b), with data points from different sources spread out in the whole encoding space.
A small number of samples (352 out of 47,393; 0.7\%) with abnormally large embedding norms, caused by numerical instability in the covariance inversion for the reference source, were excluded from the CORAL-corrected visualization.

\clearpage
\section{: MoCoP Training Hyperparameters}
\label{app:mocop_hparams}

Table~\ref{tab:mocop_hparams} summarizes the key hyperparameters used for \mocop{} v2 contrastive pretraining.

\begin{table}[htbp]
  \caption{\mocop{} v2 contrastive pretraining hyperparameters.}
  \label{tab:mocop_hparams}
  \centering
  \begin{tabular}{ll}
    \toprule
    Hyperparameter & Value \\
    \midrule
    \multicolumn{2}{l}{\textit{Molecular encoder (GGNN)}} \\
    Input atom features & 75 \\
    Edge types & 1 \\
    Message-passing layers & 6 \\
    FC dimensions & 1024 \\
    Dropout & 0.1 \\
    \midrule
    \multicolumn{2}{l}{\textit{Morphology encoder (MLP)}} \\
    Input dimension & 128 (DL embedding) \\
    Hidden dimensions & [512, 256, 128] \\
    Dropout & 0.1 \\
    \midrule
    \multicolumn{2}{l}{\textit{Projection and objective}} \\
    Shared embedding dimension & 128 \\
    Projection activation & ReLU (before linear projection) \\
    \midrule
    \multicolumn{2}{l}{\textit{Optimization}} \\
    Batch size & 256 \\
    Optimizer & AdamW \\
    Learning rate & $10^{-3}$ \\
    Scheduler & Cosine annealing with warm restarts \\
    First cycle steps & 1,000 \\
    Warmup steps & 50 \\
    Min learning rate & $10^{-8}$ \\
    Max epochs & 1,000 \\
    Early stopping metric & validation recall accuracy \\
    Early stopping patience & 500 \\
    Molecular ID & canonical SMILES \\
    Pad length & 250 \\
    \bottomrule
  \end{tabular}
\end{table}

\clearpage
\section{: Additional MoCoP Embedding Divergence Examples}
\label{app:divergence_examples}

This appendix presents additional examples where \mocop{} v1 and v2 embeddings disagree on compound similarity, beyond those shown in the main text.
All compound pairs are from Source~3 (same microscope) to eliminate batch effects.
In all cases, visual inspection of Cell Painting images supports the v2 assessment.
Cell Painting composite: Blue = DNA, Green = AGP, and Red = Mitochondria.
Plate barcodes and well positions from JUMP-CP are shown along with the structures.

\subsubsection{Additional strong divergence examples.}

\begin{figure}[htbp]
  \centering
  \begin{subfigure}[t]{0.48\textwidth}
    \centering
    \includegraphics[width=\textwidth]{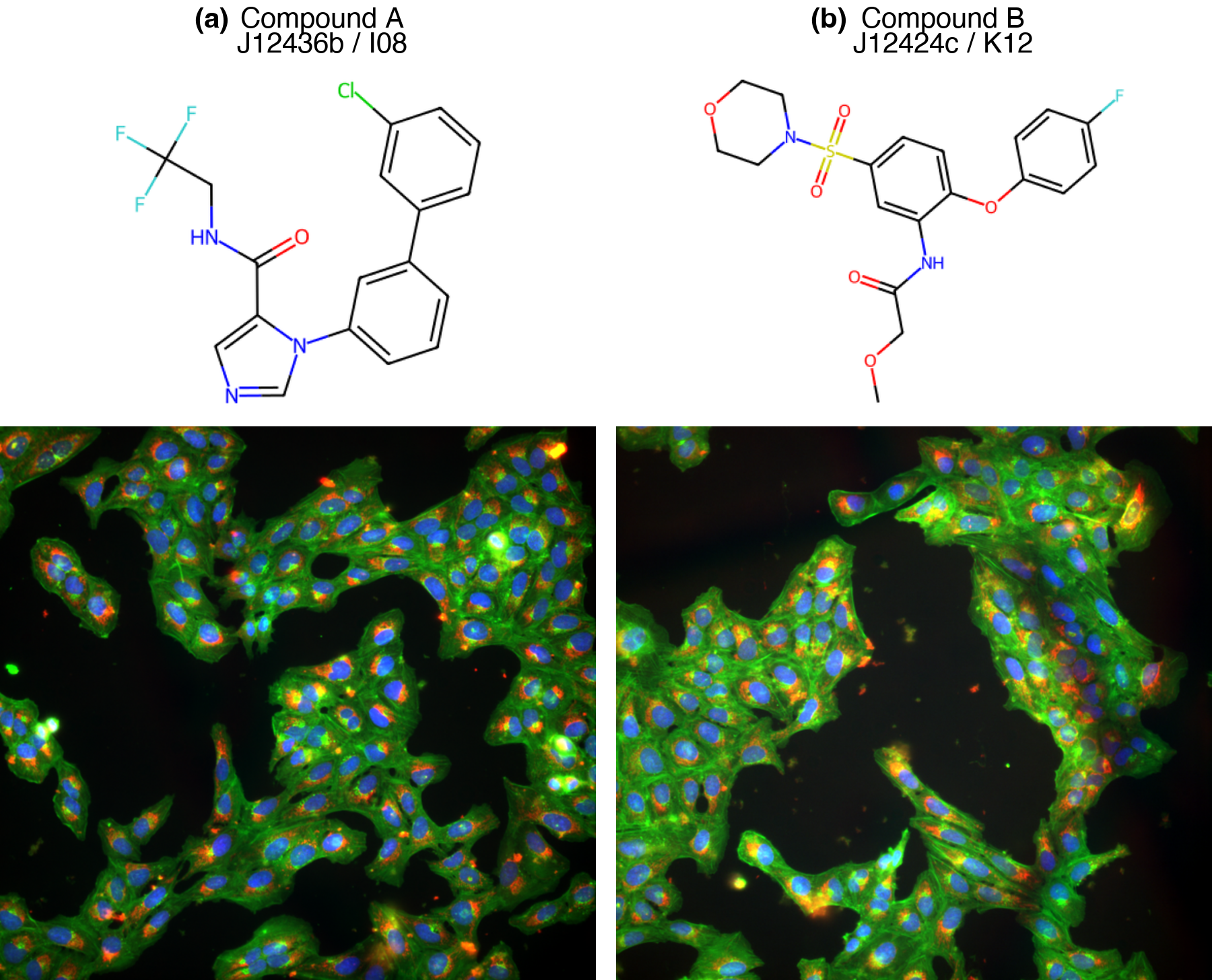}
    \caption{v2 cos $= +0.583$, v1 cos $= -0.308$, Tanimoto $= 0.135$.
    Both compounds produce similar cell density and spreading patterns with concordant channel balance.}
  \end{subfigure}
  \hfill
  \begin{subfigure}[t]{0.48\textwidth}
    \centering
    \includegraphics[width=\textwidth]{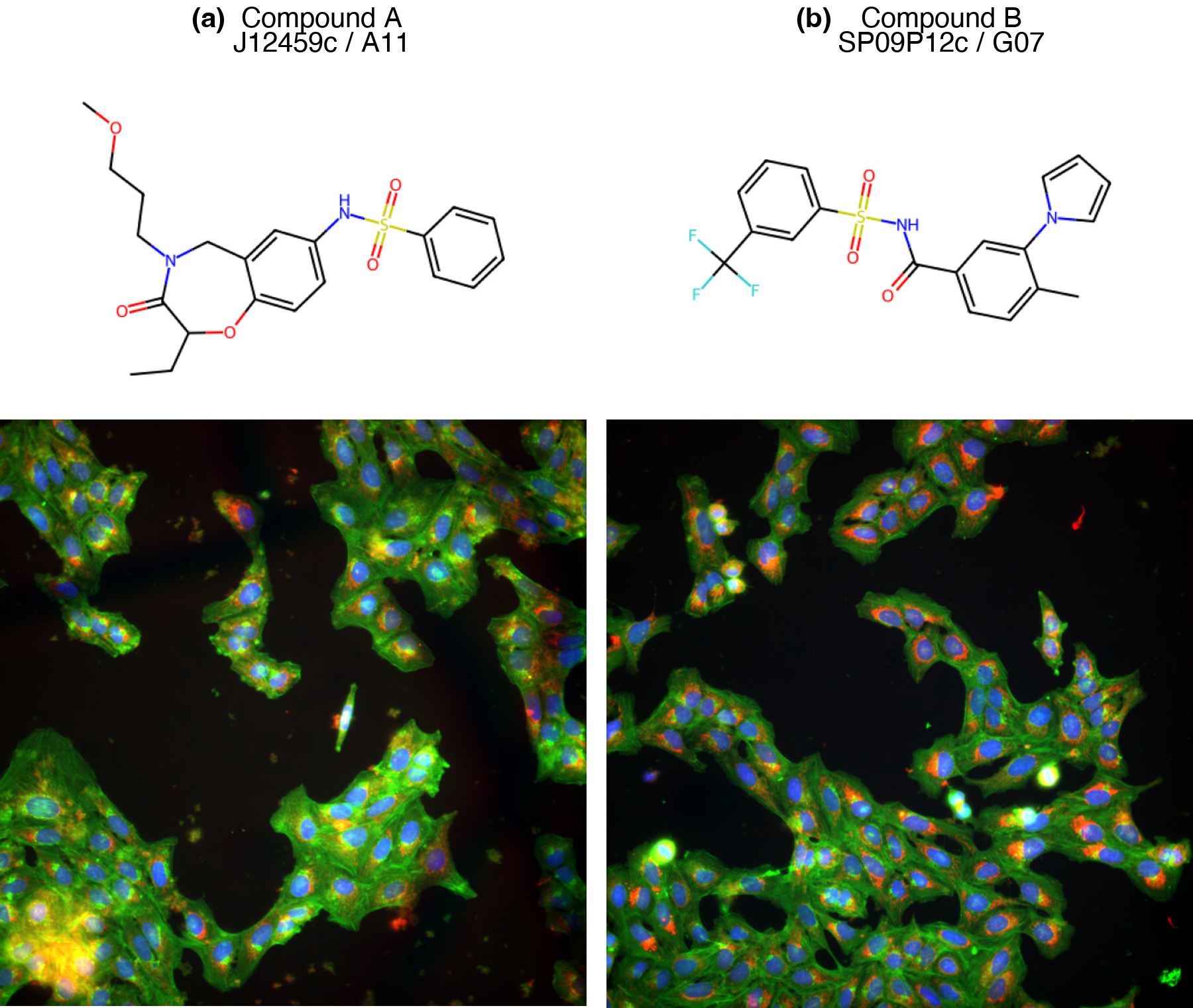}
    \caption{v2 cos $= +0.567$, v1 cos $= -0.316$, Tanimoto $= 0.170$.
    Concordant phenotypes with similar cell morphology and AGP/Mito staining patterns despite low structural similarity.}
  \end{subfigure}
  \caption{Additional strong divergence: v2-similar pairs. \mocop{} v2 correctly identifies phenotypic similarity not captured by v1.}
  \label{fig:app_divergence_v2sim}
\end{figure}

\begin{figure}[htbp]
  \centering
  \begin{subfigure}[t]{0.48\textwidth}
    \centering
    \includegraphics[width=\textwidth]{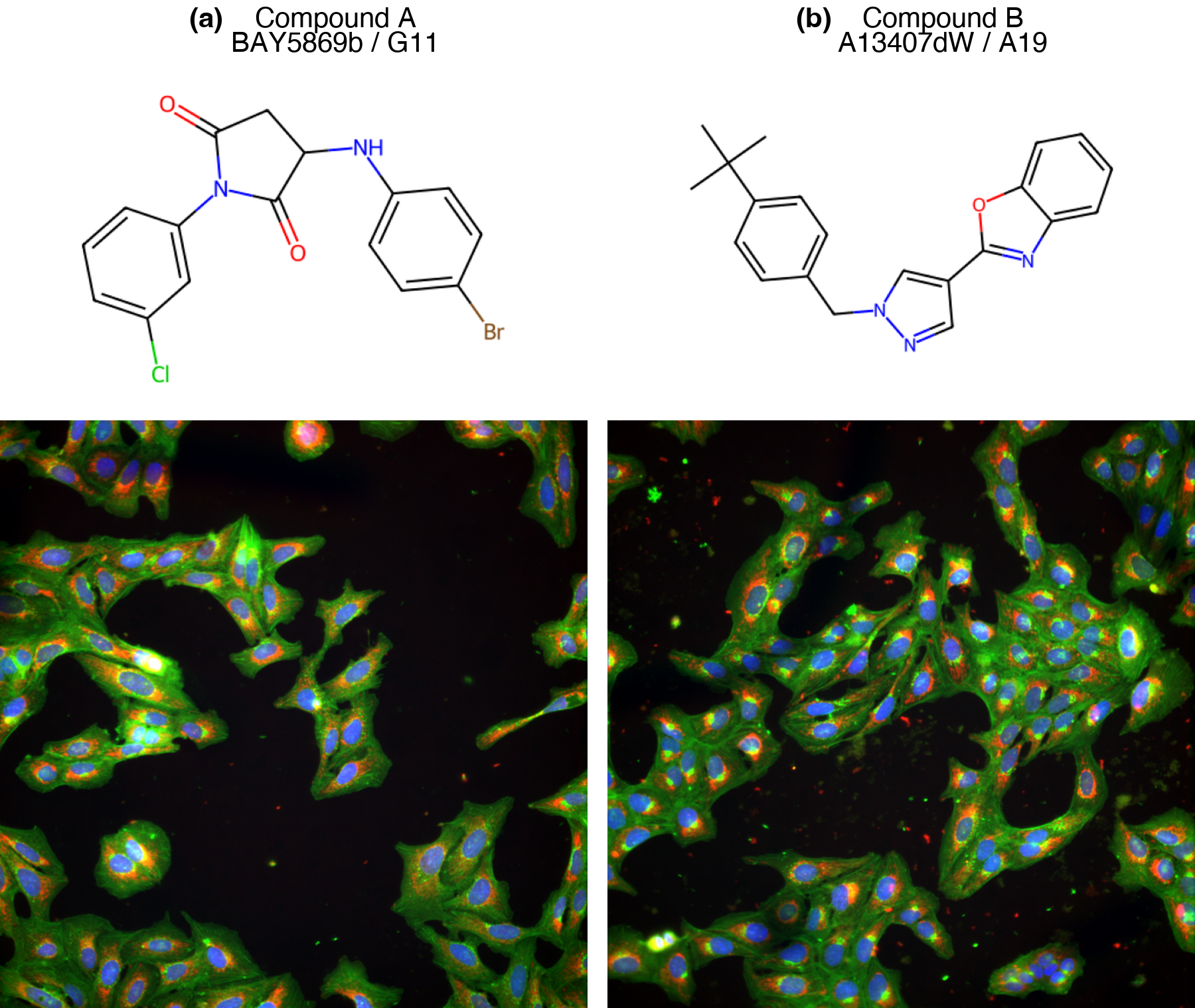}
    \caption{v1 cos $= +0.513$, v2 cos $= -0.459$, Tanimoto $= 0.107$.
    Compound~A shows large, spread cells; Compound~B shows compact, smaller cells with different morphology.}
  \end{subfigure}
  \hfill
  \begin{subfigure}[t]{0.48\textwidth}
    \centering
    \includegraphics[width=\textwidth]{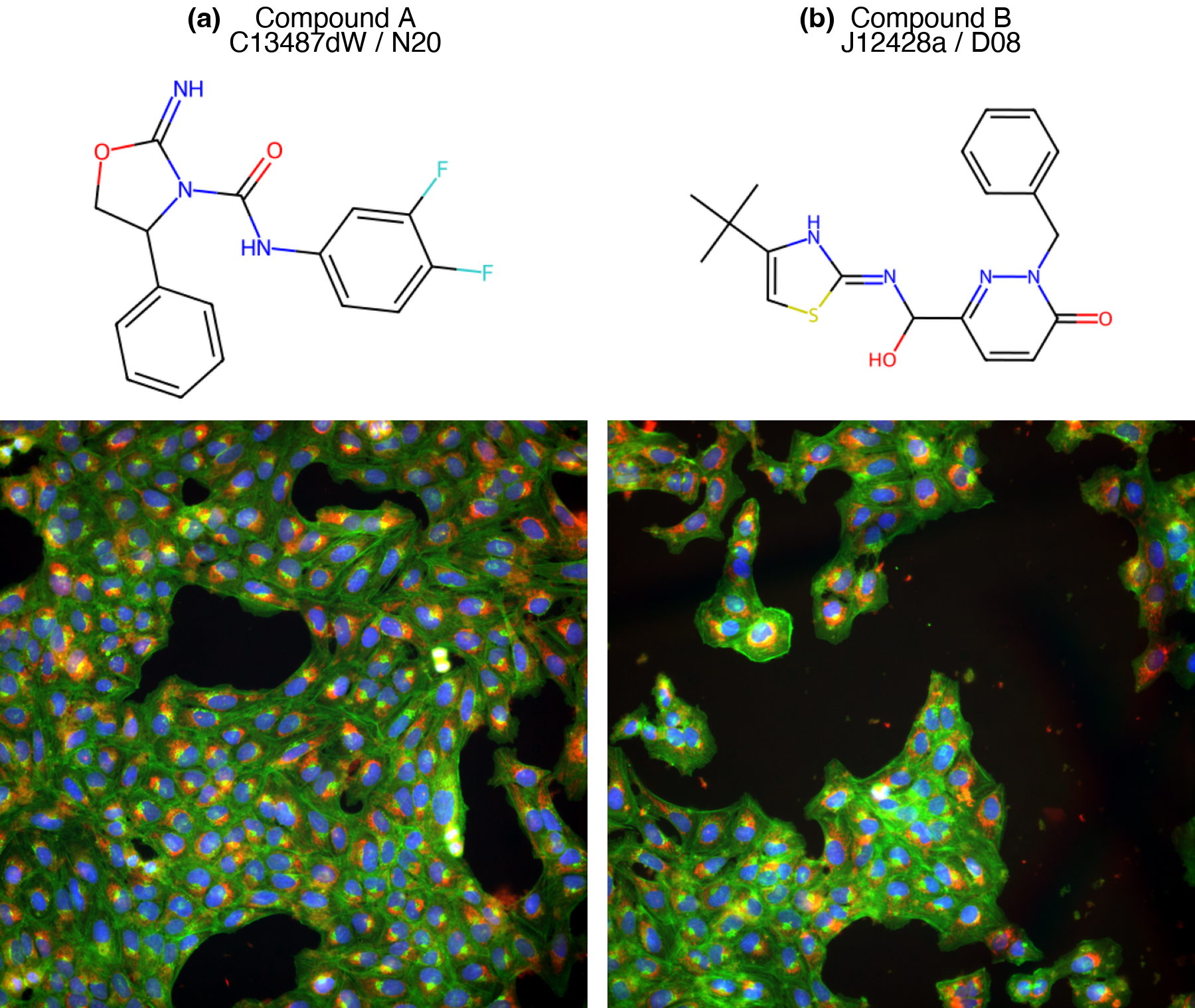}
    \caption{v1 cos $= +0.626$, v2 cos $= -0.308$, Tanimoto $= 0.126$.
    Compound~A shows very dense, small cells at high confluence; Compound~B shows larger and more sparse cells with lower occupation of the FOV, indicating different viability.}
  \end{subfigure}
  \caption{Additional strong divergence: v1-similar pairs. \mocop{} v1 incorrectly assigns high similarity (cos $> 0.5$) to compound pairs with very different phenotypes.
  v2 correctly identifies these as dissimilar.}
  \label{fig:app_divergence_v1sim}
\end{figure}

Figure~\ref{fig:app_divergence_v2sim} shows pairs where v2 correctly identifies phenotypic similarity that v1 misses.
Figure~\ref{fig:app_divergence_v1sim} shows pairs where v1 incorrectly assigns high similarity to compounds with different phenotypes.

\subsubsection{Divergence examples among structurally dissimilar compounds.}

\begin{figure}[htbp]
  \centering
  \begin{subfigure}[t]{0.48\textwidth}
    \centering
    \includegraphics[width=\textwidth]{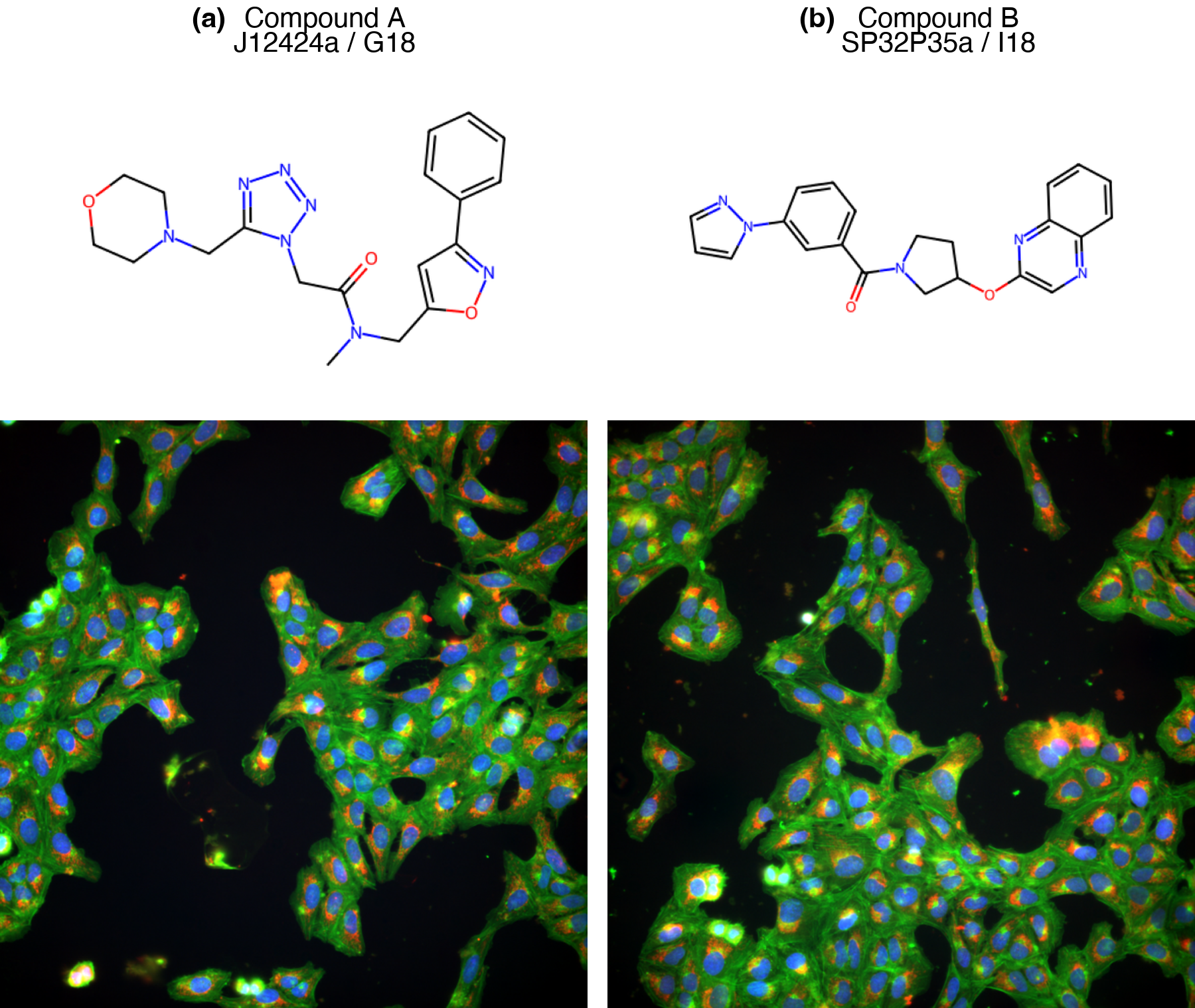}
    \caption{V2-similar, example 1 (Tanimoto $= 0.152$, v2 cos $= +0.789$, v1 cos $= +0.254$).
    Both compounds show similar cell density and spreading with concordant channel proportions.}
  \end{subfigure}
  \hfill
  \begin{subfigure}[t]{0.48\textwidth}
    \centering
    \includegraphics[width=\textwidth]{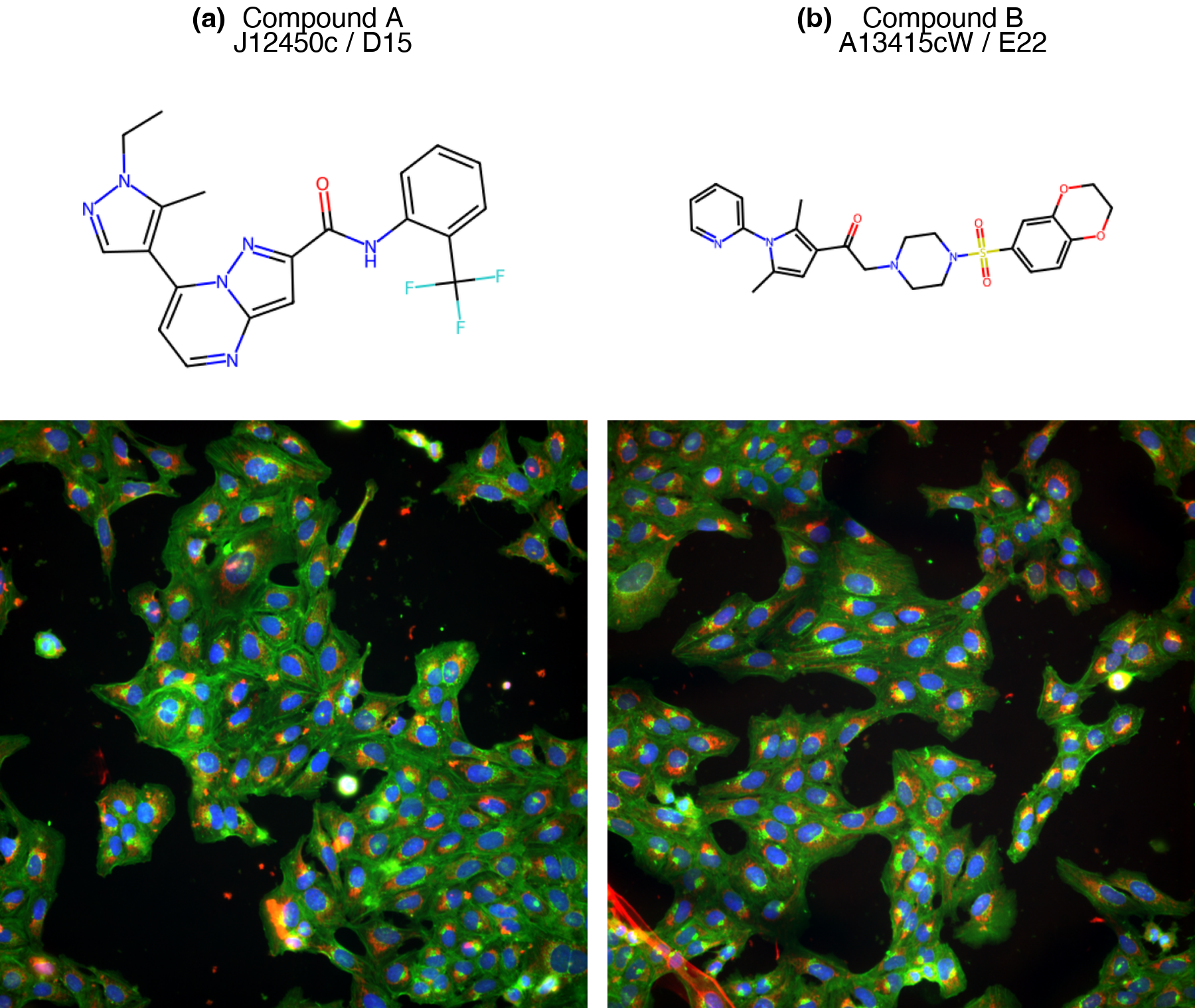}
    \caption{V2-similar, example 2 (Tanimoto $= 0.172$, v2 cos $= +0.800$, v1 cos $= +0.257$).
    Concordant morphology with similar cell size and distribution patterns.}
  \end{subfigure}
  \\[1em]
  \begin{subfigure}[t]{0.48\textwidth}
    \centering
    \includegraphics[width=\textwidth]{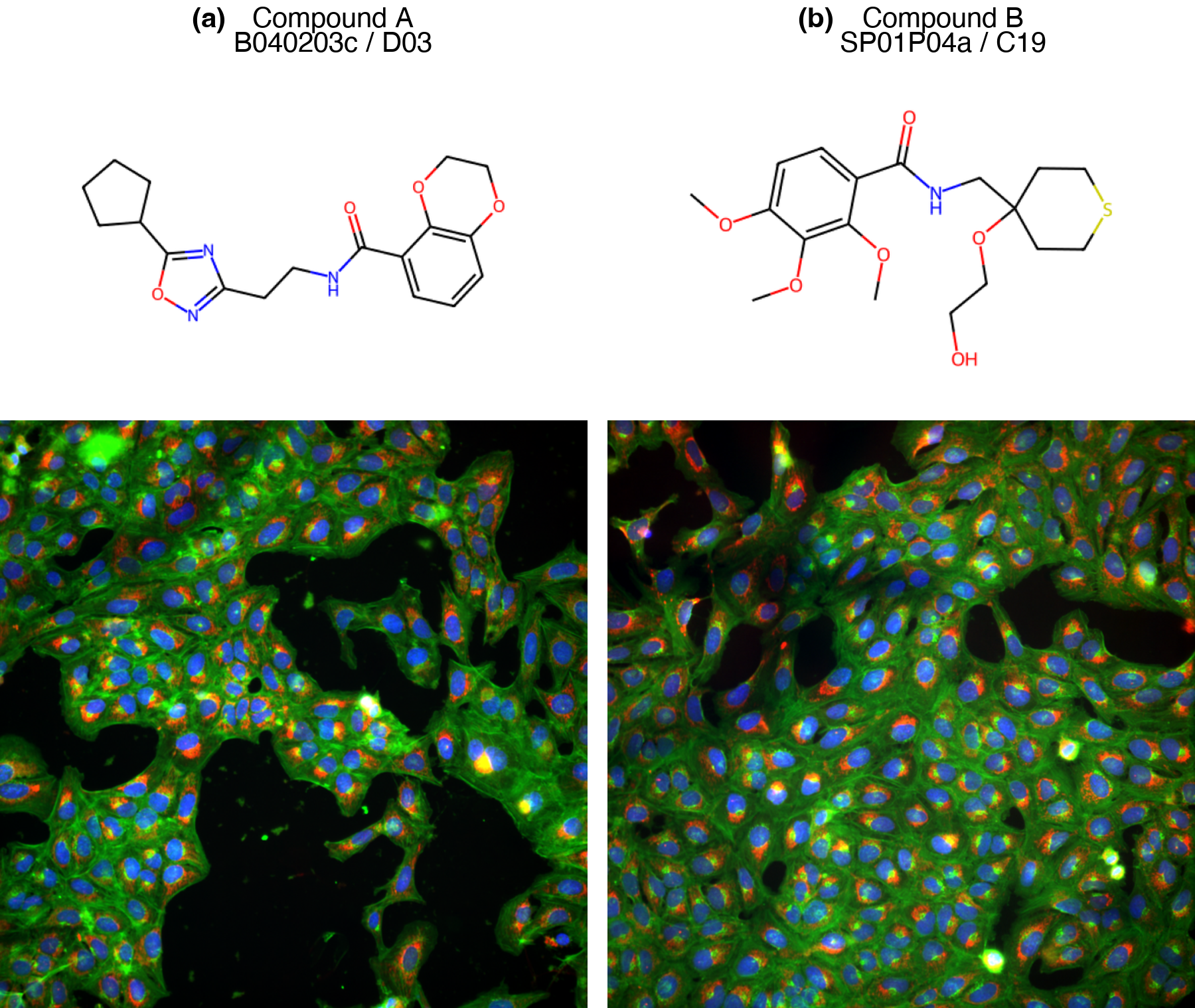}
    \caption{V1-similar, example 1 (Tanimoto $= 0.161$, v1 cos $= +0.835$, v2 cos $= +0.184$).
    Compound~A shows sparse, large cells with irregular morphology; Compound~B shows dense, uniformly distributed smaller cells. v1 incorrectly rates these as similar.}
  \end{subfigure}
  \hfill
  \begin{subfigure}[t]{0.48\textwidth}
    \centering
    \includegraphics[width=\textwidth]{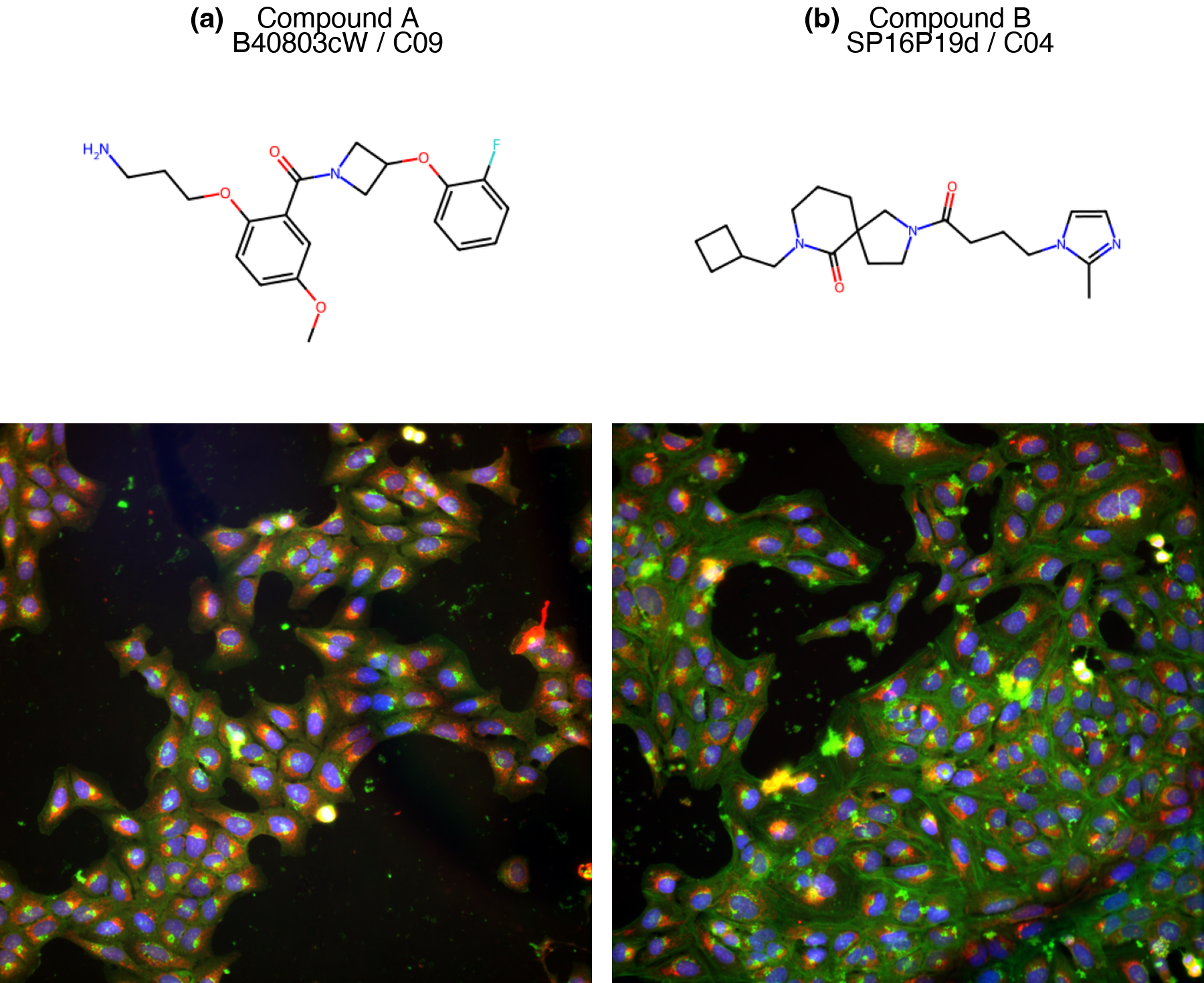}
    \caption{V1-similar, example 2 (Tanimoto $= 0.160$, v1 cos $= +0.812$, v2 cos $= +0.139$).
    Compound~A shows sparse cells with bright mitochondrial staining; Compound~B shows a denser field with different morphology and staining balance.}
  \end{subfigure}
  \caption{Divergence examples among structurally dissimilar compounds (Tanimoto $< 0.2$), imaged on the same microscope (Source~3).
  \textbf{(a--b)} V2-similar: v2 correctly identifies phenotypic similarity that v1 misses, demonstrating the value of DL-embedding-aligned representations for identifying shared biology across distinct scaffolds.
  \textbf{(c--d)} V1-similar: v1 incorrectly assigns high similarity to pairs with visibly different phenotypes; v2's lower score is more appropriate.}
  \label{fig:moderate_divergence}
\end{figure}

To further investigate the behavior of v1 and v2 for chemically isolated compounds, we selected pairs with low Tanimoto similarity ($< 0.2$) and moderate embedding divergence between the two models.
Figure~\ref{fig:moderate_divergence} shows two types of disagreement:

\begin{itemize}
  \item \textbf{V2-similar pairs} (Figure~\ref{fig:moderate_divergence}a--b): v2 cosine $\approx +0.80$ (95th percentile), v1 cosine $\approx +0.25$ (64th percentile).
  Visual inspection confirms concordant phenotypes (similar cell density, spreading pattern, and channel intensity balance) despite very different chemical structures (Tanimoto $\approx 0.15$).
  \mocop{} v2 correctly identifies this morphological similarity, demonstrating its utility for finding phenotypically related compounds across distinct scaffolds.

  \item \textbf{V1-similar pairs} (Figure~\ref{fig:moderate_divergence}c--d): v1 cosine $\approx +0.83$ (96th percentile), v2 cosine $\approx +0.18$ (61st percentile).
  Visual inspection reveals visibly different phenotypes with clear differences in cell density, size, morphology, and staining patterns.
  \mocop{} v1 incorrectly assigns high similarity; v2's lower score is more appropriate.
\end{itemize}

\clearpage
\section{: Downstream Task Training Details}
\label{app:downstream_details}

For all downstream benchmark tasks, we followed the evaluation protocol established by InfoAlign~\cite{Liu2024InfoAlign}.
Datasets were obtained from the InfoAlign repository, with scaffold-based splitting used for train/validation/test partitioning.
Classification tasks used AUROC as the primary metric; regression tasks (Biogen~3K) used MAE.
All results are reported as 3-seed mean $\pm$ standard deviation on the test set, by training separate models initialized from the pretrained checkpoint.

Table~\ref{tab:downstream_datasets} summarizes the benchmark datasets.

\begin{table}[htbp]
  \caption{Summary of downstream benchmark datasets.}
  \label{tab:downstream_datasets}
  \centering
  \begin{tabular}{lccc}
    \toprule
    Dataset & Task Type & \# Compounds & \# Tasks \\
    \midrule
    ChEMBL~2K\cite{chembl2k} & Classification (AUC) & 2,355 & 41 \\
    ToxCast\cite{toxcast} & Classification (AUC) & 8,576 & 617 \\
    Broad~6K\cite{Becker2020PredictingActivity} & Classification (AUC) & 6,567 & 32 \\
    Biogen~3K\cite{biogen3k} & Regression (MAE) & 3,521 & 6 \\
    \bottomrule
  \end{tabular}
\end{table}

\subsubsection{Hyperparameter Sweep.}
A comprehensive hyperparameter sweep was conducted across learning rate ($5 \times 10^{-5}$ to $10^{-3}$), dropout (0.1--0.3), early stopping patience (10--20 epochs), optimizer (Adam vs.\ AdamW), and learning rate scheduler (constant vs.\ cosine annealing).
Key findings from the sweep:
\begin{itemize}
  \item Higher learning rates (3$\times 10^{-4}$ to 7$\times 10^{-4}$) consistently improved validation performance over the baseline ($5 \times 10^{-5}$).
  \item Reduced early stopping patience (10 vs.\ 20 epochs) reduced overfitting to the validation set.
  \item Moderate dropout (0.2) generalized better than 0.1 at high learning rates; 0.3 hurt validation performance.
  \item Cosine annealing provided marginal improvement for ChEMBL~2K only (+0.3\% AUC).
  \item AdamW weight decay did not outperform Adam.
\end{itemize}

Table~\ref{tab:best_configs} reports the best configuration for each dataset.

\begin{table}[htbp]
  \caption{Best hyperparameter configurations for each downstream dataset.}
  \label{tab:best_configs}
  \centering
  \begin{tabular}{lcccc}
    \toprule
    Dataset & Learning Rate & Dropout & Patience & Scheduler \\
    \midrule
    ToxCast & $3 \times 10^{-4}$ & 0.2 & 10 & --- \\
    ChEMBL~2K & $5 \times 10^{-4}$ & 0.2 & 10 & Cosine ($T=50$) \\
    Broad~6K & $5 \times 10^{-5}$ & 0.1 & 20 & --- \\
    Biogen~3K & $7 \times 10^{-4}$ & 0.2 & 10 & --- \\
    \bottomrule
  \end{tabular}
\end{table}

\end{document}